\documentclass[11pt]{article}

\usepackage[preprint]{acl}
\usepackage{times}
\usepackage{latexsym}
\usepackage[T1]{fontenc}
\usepackage[utf8]{inputenc}
\usepackage{microtype}
\usepackage{inconsolata}
\usepackage{graphicx}
\usepackage{booktabs}
\usepackage{amsmath}
\usepackage{amssymb}
\usepackage{array}
\usepackage{tabularx}
\usepackage{placeins}
\usepackage{float}
\usepackage{enumitem}

\graphicspath{{figures/}{figures/supplement/}}
\newcommand{\method}{\textsc{BasinLens}}
\newcommand{\frontier}{\textsc{BasinLens-Frontier}}
\newcommand{\interaction}{\textsc{BasinLens-Interaction}}
\newcommand{\beam}{\textsc{BasinLens-Beam}}

\title{Where World Models Break: Natural-Input Failure Discovery}
\author{
    \normalsize
    \textbf{Zhanpeng Shi}\textsuperscript{1,\textdagger},
    \textbf{Zi Liang}\textsuperscript{2,\textdagger},
    \textbf{Rong Feng}\textsuperscript{3},\\
    \textbf{Shiqin Tang}\textsuperscript{4},
    \textbf{Xuyang Chen}\textsuperscript{5},
    \textbf{Hongzong Li}\textsuperscript{1,6,*}\\
    {\small \textsuperscript{1}School of Computer Science, Northwestern Polytechnical University}\\[-2pt]
    {\small \textsuperscript{2}The Hong Kong Polytechnic University}\\[-2pt]
    {\small \textsuperscript{3}Department of Data Science, City University of Hong Kong}\\[-2pt]
    {\small \textsuperscript{4}Centre for Artificial Intelligence and Robotics, Hong Kong Institute of Science \& Innovation,}\\[-2pt]
    {\small Chinese Academy of Sciences}\\[-2pt]
    {\small \textsuperscript{5}Department of Electrical and Computer Engineering, National University of Singapore}\\[-2pt]
    {\small \textsuperscript{6}Generative AI Research and Development Center, The Hong Kong University of Science and Technology}\\[-2pt]
    {\small \textsuperscript{\textdagger}Equal contribution.\quad
    \textsuperscript{*}Corresponding author.}\\[-2pt]
    {\small \textbf{Correspondence:} \texttt{lihongzong@nwpu.edu.cn}}
}

\begin{document}
\maketitle


\begin{abstract}
World models predict action-conditioned futures and serve as critical internal simulators for downstream planning and control. However, catastrophic prediction failures of world models could dangerously propagate through the control pipeline, as subsequent agent or model training and decision-making depend heavily on the continuous environment evolution forecasted by these world models. Existing evaluations overlook this systemic risk: by aggregating average errors over benign generations from general queries, they fail to stress-test the model against catastrophic collapses under rare or unobserved condition-action combinations.
To bridge this gap, we formalize the \emph{natural-input failure discovery} problem: under a finite query budget, finding environment-valid conditions and action prefixes that induce severe prediction risk, verifying whether these failures reproduce on fresh seeds, and testing their persistence under nearby valid edits. Discovering such critical failures is computationally challenging, as valid condition-action combinations explode exponentially, rendering exhaustive search or standard sampling infeasible given the high cost of noisy rollouts. To tackle this, we propose \method{}, which exploits the underlying structure of valid inputs, where each coordinate possesses environment-defined semantic types and admissible domains, by pairing uncertainty-guided global search with typed local replacements. Across diverse benchmarks and world-model families, \method{} exposes reproducible and locally persistent failure modes that conventional evaluations fail to reveal, showing that average-case benchmarks can mask important vulnerabilities in world-model-driven control.
\end{abstract}

\section{Introduction}
World models learn action-conditioned environment dynamics in observation space or in task-relevant latent representations. They now serve as internal simulators for model-based planning and control, from recurrent imagination models \cite{ha2018worldmodels,hafner2020dreamer} to scalable latent-control systems \cite{hafner2023dreamerv3,hansen2024tdmpc2}. Because planners compare futures predicted by the model, a localized forecast error can change action rankings even when aggregate prediction error or benchmark return remains strong. Evaluation should therefore ask not only how well a world model performs on average, but which environment-valid resets or action sequences make its forecasts unreliable. WorldBench reports aggregate and per-concept fidelity over randomized video continuations \cite{upadhyay2026worldbench}, while interactive and closed-loop evaluations aggregate behavior over sampled interactions or trajectories \cite{kong2026worldlens,li2026dworldeval}. These protocols compare systems but do not adaptively identify valid conditions that repeatedly trigger local failures.

Robustness methods address related but different targets. Adversarial examples perturb inputs \cite{goodfellow2015adversarial}, adversarial policies attack victims through legal opponent actions \cite{gleave2020adversarialpolicies}, and world-model attacks perturb physical-conditioning channels \cite{guo2026physcondwma}. These methods target classifiers, policies, or features, not fixed-model prediction failures over valid inputs. Adaptive stress testing targets decision failures \cite{koren2018ast}, while VerifAI searches Scenic-defined scenarios for specification violations \cite{dreossi2019verifai,fremont2019scenic}. Gaussian-process Bayesian optimization can locate high-loss inputs \cite{srinivas2010gpucb}, but it does not determine whether a discovered failure reproduces across seeds or persists under nearby legal changes. We therefore formalize \emph{natural-input failure discovery}: given a fixed world model, a finite query budget, and an environment-defined valid-input set, locate reset conditions or action prefixes that induce high prediction risk. The difficulty is twofold. The valid input space grows combinatorially, and evaluating each candidate requires costly and noisy model and environment rollouts. Because adaptive search favors extreme observations, selected failures must be tested independently for reproducibility and local persistence.

\begin{figure*}[t]
\centering
\includegraphics[width=.98\textwidth]{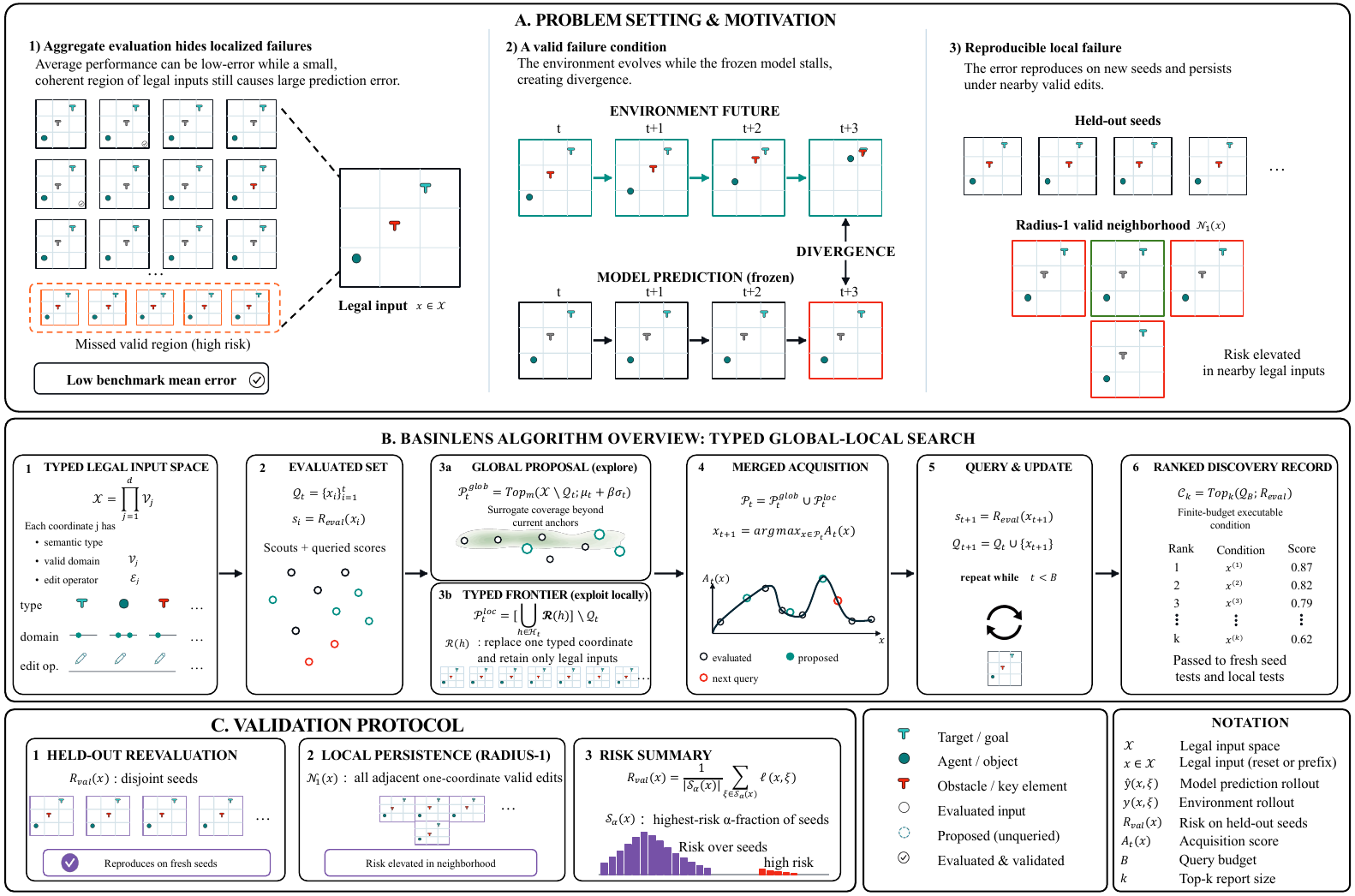}
\caption{Problem, structured search, and independent evidence. Aggregate evaluation can miss a small region of legal inputs (left). \method{} represents legal conditions as typed coordinates, maintains the evaluated set, forms global-surrogate and typed-frontier proposal pools, and queries their merged acquisition until returning a finite-budget discovery record (center). Held-out-seed re-evaluation and fresh radius-1 legal neighborhoods test reproducibility and local persistence without feeding validation evidence back into search (right).}
\label{fig:failure-discovery}
\end{figure*}

Valid inputs have explicit coordinate structure. Each input coordinate has a semantic type, an admissible domain, and an environment-valid edit operator. Replacing one coordinate with another admissible value yields an executable and interpretable local move without assuming that all normalized directions have the same semantics. \method{} combines uncertainty-guided global proposals with typed local replacements (Figure~\ref{fig:failure-discovery}). Global proposals maintain coverage, typed replacements refine high-risk candidates, and a joint acquisition rule allocates the remaining budget. The search returns a ranked set of concrete inputs. After the search trace is fixed, these inputs are evaluated on held-out seeds and in fresh legal neighborhoods without updating the acquisition process. In this framing, \method{} is one implementation of the discovery problem, while the independently checkable failure record is the primary scientific object.

On a 12,016-candidate online PushT grid \cite{chi2023diffusionpolicy,florence2022implicit}, fixed-threshold discovery at budget 64 occurs in $54.7\%$ of \method{} runs, $40.6\%$ of Gaussian-process upper confidence bound (GP-UCB) runs, and $10.9\%$ of random-search runs; the GP-UCB comparison at this threshold is descriptive. Selected inputs retain elevated risk on disjoint seeds, and radius-1 neighborhoods around top-score anchors reach mean conditional value at risk (CVaR) $1.252$, compared with at most $0.608$ for controls. Studies on four additional world-model families show that the discovery formulation transfers, while also revealing a useful boundary: typed expansion is most helpful in heterogeneous, interacting input spaces, whereas standard search can remain preferable in low-dimensional or homogeneous spaces.

Our contributions are threefold:
\begin{itemize}
    \item \textbf{Problem and evidence standard.} We formalize finite-budget natural-input failure discovery for fixed world models and distinguish discovery, held-out point reproducibility, and fresh-neighborhood persistence.
    \item \textbf{Structured implementation.} We introduce \method, which combines uncertainty-guided global proposals with valid typed replacements to return executable, inspectable test cases.
    \item \textbf{Empirical findings and scope.} Across five world-model families and four prediction interfaces, we expose valid failure conditions, verify selected PushT cases independently, and identify settings in which typed structure helps or standard search remains competitive.
\end{itemize}

\section{Problem Formulation}
Let $\mathcal{E}$ be an environment with observation space $\mathcal{O}$ and action space $\mathcal{A}$, let $M$ be a trained action-conditioned world model, and let $\pi$ be a fixed rollout policy. A \emph{valid natural input} $x\in\mathcal{X}$ specifies an environment-valid condition, such as reset variables, a goal, or an action prefix. An input specification defines both the admissible set $\mathcal{X}$ and which coordinate changes preserve validity. When $x$ contains an action prefix, execution applies that prefix before following $\pi$ for the remaining rollout.

For an input $x$ and environment random seed $\xi$, applying any action prefix in $x$ and then following $\pi$ in $\mathcal{E}$ produces a trajectory
\[
\tau(x,\xi)=\bigl(o_0,a_0,o_1,a_1,\ldots,a_{T-1},o_T\bigr),
\]
where $o_t\in\mathcal{O}$ is the observation at time $t$, $a_t\in\mathcal{A}$ is the action applied after observing $o_t$, and $T$ is the rollout length. At a forecast origin $H$ and prediction horizon $K$, define the observed history as $h_H=(o_0,a_0,\ldots,a_{H-1},o_H)$ and the future action sequence as $a_{H:H+K-1}=(a_H,\ldots,a_{H+K-1})$. The world model predicts
\[
\widehat{y}_{H+1:H+K}=M\bigl(h_H,a_{H:H+K-1}\bigr).
\]
The corresponding reference target is
\[
y_{H+1:H+K}=\Phi\bigl(o_{H+1:H+K}\bigr),
\]
where $K\geq1$, $0\leq H<H+K\leq T$, model-side decoding or readout is absorbed into $M$, and $\Phi$ maps the environment future into the same evaluation space. Depending on the interface, $\Phi$ can be an encoder, a simulator-state readout, or the identity map in observable space.

Let $d(\cdot,\cdot)$ be a nonnegative discrepancy function, such as mean squared error in a latent or decoded space. The per-rollout prediction risk is
\[
\ell(x,\xi)=d\!\left(\widehat{y}_{H+1:H+K},y_{H+1:H+K}\right).
\]
For an evaluation seed set $\mathcal{S}_{\mathrm{eval}}$, the aggregate prediction risk of $x$ is
\[
R_{\mathrm{pred}}(x)=\rho\bigl(\{\ell(x,\xi):\xi\in\mathcal{S}_{\mathrm{eval}}\}\bigr),
\]
where $\rho$ is a prespecified aggregation operator, such as the mean, maximum, or upper-tail conditional value at risk (CVaR) \cite{rockafellar2000cvar}. When a protocol also requires task-active rollouts, let $v(x)\geq0$ be its prespecified activity violation and define the search score
\[
R_{\mathrm{eval}}(x)=R_{\mathrm{pred}}(x)-\lambda_{\mathrm{act}}v(x),
\]
where $\lambda_{\mathrm{act}}\geq0$ is the task-activity penalty weight; protocols without this adjustment set $\lambda_{\mathrm{act}}=0$. An adaptive algorithm may evaluate at most $B$ valid inputs, producing $\mathcal{Q}_B\subseteq\mathcal{X}$ with $|\mathcal{Q}_B|\leq B$. For a prespecified shortlist size $1\leq k\leq|\mathcal{Q}_B|$, it selects
\[
\mathcal{C}_k
=\operatorname{Top}_k\!\left(\mathcal{Q}_B;R_{\mathrm{eval}}\right)
\]
and, when fresh evaluation is available, returns validation records
\[
\widehat{\mathcal{V}}_k
=\{(x,R_{\mathrm{val}}(x)):x\in\mathcal{C}_k\},
\]
where $\operatorname{Top}_k(\mathcal{Q};R)$ returns the $k$ elements of $\mathcal{Q}$ with the largest values under $R$ and a fixed tie-breaking rule, and $R_{\mathrm{val}}$ uses a disjoint seed set $\mathcal{S}_{\mathrm{val}}$. Only shortlisted candidates, rather than every queried input, receive held-out validation. Search performance can be measured by best discovered score, the probability of finding any input above a threshold $\theta$, or threshold-averaged discovery. The goal is to identify a small set of valid inputs for inspection, not to estimate average risk under a deployment distribution.

For validation, the input specification induces a valid-input graph $\mathcal{G}=(\mathcal{X},\mathcal{E}_{\mathcal{G}})$, where $(x,x')\in\mathcal{E}_{\mathcal{G}}$ when one valid edit moves one typed coordinate to an adjacent value in its ordered grid. Let $d_{\mathcal{G}}$ be the shortest-path distance on this graph. For radius $r\in\mathbb{N}_0$, define
\[
\mathcal{N}_r(x)=\{x'\in\mathcal{X}:d_{\mathcal{G}}(x,x')\leq r\}
\]
as the radius-$r$ valid neighborhood. Its fresh-evaluation mean risk is
\[
\overline{R}_{\mathrm{val},r}(x)
=\frac{1}{|\mathcal{N}_r(x)|}
\sum_{x'\in\mathcal{N}_r(x)}R_{\mathrm{val}}(x').
\]
We compare this quantity with neighborhoods around prespecified control anchors. Cached diagnostic studies additionally use a broader replacement graph, connecting inputs that differ by any valid single-coordinate replacement, and measure how many oracle top-score inputs from one reference component are covered within budget. This reference-component metric is defined in the supplement and is not part of the core failure definition.

Validity does not imply prevalence. A natural input is executable through the environment's ordinary reset or action interface and passes prespecified validity checks; it need not be common under an unknown deployment distribution. The finite library defines a reproducible search domain and is not used to estimate failure frequency. Discovery returns concrete valid conditions under a stated query budget. Each reported score is specific to its prediction interface, evaluation seeds, and validity rules.

\section{A Structured Implementation: \method}
\method{} instantiates natural-input failure discovery as a structured budgeted search (Figure~\ref{fig:failure-discovery}). It maintains an evaluated set, forms complementary global and typed-local candidate pools, ranks their union, and returns a shortlist for independent follow-up evaluation. A global surrogate can reach distant candidates, but normalized geometry alone does not identify meaningful or environment-valid edits. A frontier preserves typed semantics, but a frontier-only search can remain confined around early anchors. \method{} therefore uses global proposals for coverage and typed replacements for executable local refinement; held-out and adjacency-neighborhood measurements remain separate from acquisition.

\subsection{Typed Valid Frontier}
Let $\mathcal{Q}_t$ be the inputs evaluated after $t$ queries and let $\mathcal{H}_t\subseteq\mathcal{Q}_t$ contain the highest-scoring evaluated inputs used as anchors. Let the \emph{replacement neighborhood} $\mathcal{R}(x)$ contain valid candidates obtained by replacing one typed coordinate of $x$ with any other allowed value. The unevaluated local frontier is
\[
\mathcal{P}^{\mathrm{loc}}_t
=\left(\bigcup_{x\in\mathcal{H}_t}\mathcal{R}(x)\right)
\setminus\mathcal{Q}_t.
\]
An edit substitutes an admissible value for one typed reset field or action-prefix entry; the input specification retains environment-valid candidates. Search uses the broader replacement neighborhood for exploration, whereas persistence validation uses the stricter \emph{adjacency neighborhood} $\mathcal{N}_1$, which changes one coordinate to an adjacent grid value. The coordinate-only variant \frontier{} uses the replacement frontier without global proposals. This mechanism is related to combinatorial testing over parameter-value combinations \cite{kuhn2013combinatorial}, but operates directly on valid environment variables.

\subsection{Global--Local Acquisition}
\method{} augments the frontier with global exploration. A normalized representation $\psi(x)$ maps each typed input to a coordinate scale. After every evaluation, the method fits a Gaussian-process surrogate to $\{(\psi(x),R_{\mathrm{eval}}(x)):x\in\mathcal{Q}_t\}$. For an unevaluated input $x$, the surrogate provides a posterior mean $\mu_t(x)$ and standard deviation $\sigma_t(x)$. Following GP-UCB \cite{srinivas2010gpucb}, its score uses a fixed standard-deviation multiplier:
\[
a_{\mathrm{UCB},t}(x)=\mu_t(x)+\beta\sigma_t(x),
\]
where $\beta>0$ controls exploration. The global pool $\mathcal{P}^{\mathrm{glob}}_t$ contains the highest-scoring unevaluated inputs under $a_{\mathrm{UCB},t}$. Unlike frontier expansion, this pool can propose candidates far from the current anchors.

The two candidate sources are combined as
\[
\mathcal{P}_t=\mathcal{P}^{\mathrm{loc}}_t\cup\mathcal{P}^{\mathrm{glob}}_t.
\]
Within this pool, a practical ranking rule standardizes the UCB score by the mean and standard deviation of the observed search scores and augments it with three structural terms:
\begin{align}
A_t(x)={}&z^{\mathrm{UCB}}_t(x)
+ \lambda_p z^{\mathrm{par}}_t(x)\notag\\
&+ \lambda_d \kappa^{\mathrm{anc}}_t(x)
+ \lambda_n \delta^{\mathrm{nov}}_t(x).
\end{align}
Here, $z^{\mathrm{par}}_t(x)$ is the largest standardized search score among anchors whose replacements produce $x$; $\kappa^{\mathrm{anc}}_t(x)$ measures Gaussian proximity to the anchors; and $\delta^{\mathrm{nov}}_t(x)$ is the minimum normalized distance to an evaluated input. The parent and proximity terms retain local evidence, while the novelty term encourages coverage. Our experiments use one fixed setting of $\lambda_p$, $\lambda_d$, and $\lambda_n$ across all LeWM PushT online protocols and a separate model-specific setting for DINO-WM. Exact definitions and settings are reported in the supplement.

Normalization supports global surrogate modeling, whereas typed edits define valid local changes. Normalization supplies the global surrogate with a numerical coordinate system; the replacement graph defines meaningful local changes. Every edit maps back to one named field, is checked against its admissible values, and remains executable even when Euclidean proximity would be misleading. The method requires neither world-model gradients nor a differentiable simulator; it only requires a prediction-risk value for each queried input.

\subsection{Discovery Records and Independent Evidence}
When the budget is exhausted, \method{} returns the shortlist $\mathcal{C}_k$, not an inferred risk surface. Each discovery record retains the stable identifier, typed values, evaluation seeds, per-seed losses, activity measurements when applicable, ranking score, and ordered query trace. These fields make each selected condition reproducible and allow the complete search trace and comparator pairing to be verified.

Follow-up evidence is attached only after the query trace and shortlist are frozen. Held-out evaluation changes the seeds while preserving the input; neighborhood evaluation applies adjacent typed edits on fresh seeds. Neither measurement updates the surrogate or reranks the trace. The initial search identifies candidates for inspection, held-out evaluation tests pointwise reproducibility, and fresh-neighborhood evaluation tests local persistence. Together, these evaluations determine whether a discovered input represents a repeatable point failure or a locally persistent failure region.

\interaction{} adds pairwise coordinate features in combinatorial reset spaces, whereas \beam{} expands multiple high-score anchors without a global GP-UCB refresh. These cached component-isolation tools test whether interactions or multi-anchor expansion explain gains under different input structures; their complete scoring rules are reported in the supplement.

\section{Experimental Setup}
\subsection{Models and Prediction Measures}
The main study uses LeWorldModel (LeWM) checkpoints \cite{maes2026lewm} on PushT \cite{chi2023diffusionpolicy,florence2022implicit} and TwoRooms. Additional checks use DINO World Model (DINO-WM) on PushT \cite{zhou2024dinowm}, a joint-embedding predictive architecture world model (JEPA-WM) on PointMaze \cite{terver2025jepawms}, and DIAMOND and IRIS on Atari \cite{alonso2024diamond,micheli2023iris}.

The main LeWM PushT prediction risk $R_{\mathrm{pred}}$ is the upper-tail CVaR of state-probe mean squared error (MSE) at the final forecast step, computed across evaluation seeds. LeWM observes three frames and predicts three frames sampled at five-simulator-step intervals from a 40-step rollout. The tail contains the largest $\lceil0.4n\rceil$ errors among $n$ seeds: four of eight search-seed errors and seven of 16 validation-seed errors. Online acquisition subtracts a task-activity penalty from candidates that violate prespecified rollout thresholds; held-out and neighborhood risks are unpenalized CVaR. DINO-WM uses final-step visual-feature MSE, JEPA-WM uses latent visual MSE plus $0.1$ times latent proprioceptive MSE, and Atari uses future-frame MSE. Risk magnitudes are interpreted only within a protocol because prediction spaces differ across models. Exact definitions and activity penalties are in the supplement.

We evaluate LeWM through a latent-to-state probe rather than its optional diagnostic red--green--blue (RGB) decoder, obtaining PushT state-space risk and physical readouts. The LeWM results evaluate the complete pipeline from latent world-model prediction to state readout. Separate held-out checks measure the probe's in-distribution accuracy. In the displayed TwoRooms case, future-position mean absolute error (MAE) rises from $13.2$ to $89.0$ pixels while probe MAE changes from $1.13$ to $1.66$ pixels, showing that readout error alone does not explain that case. DINO-WM provides probe-independent evidence through visual-feature prediction.

\subsection{Protocols and Comparisons}
The experiments use distinct protocols for online discovery, held-out point validation, fresh-neighborhood evaluation, component isolation, and cross-model scope. Cached studies use a dense agent-start library (8,177 candidates; 3,928 environment-valid inputs) and a cached combinatorial grid (431 environment-valid combinations).

\begin{table}[t]
\centering
\small
\caption{Protocol map. Each row answers a different question; measurements are not pooled across rows.}
\label{tab:protocol-map}
\begin{tabular}{@{}lp{.23\columnwidth}p{.39\columnwidth}@{}}
\toprule
Question & Library / budget & Independent record \\
\midrule
Discovery & 12,016; $B\in\{16,32,64\}$ & Search score, fixed-threshold hit, and mean best score \\
Held-out & 576; $B=32$ & Selected points on 16 disjoint seeds \\
Local & 3,510; $B=32$ & Fresh radius-1 neighborhoods around top, mid, and random anchors \\
Components & Cached grids; other interfaces & Common-scout hit or coverage; no cross-interface scale pooling \\
\bottomrule
\end{tabular}
\end{table}

The large online grid contains 12,016 valid candidates and uses scout size three, 64 paired algorithm restarts, eight evaluation seeds per candidate, and budgets 16, 32, and 64. A separate 576-candidate point-validation library has 96 scout-size-three search runs at budget 32; its pooled validation set adds 32 existing scout-size-four runs, yielding 128 restarts, and uses 16 disjoint validation seeds. The 3,510-candidate wide validation grid couples budget-32 search with fresh radius-1 neighborhood evaluation. All three LeWM online libraries use the same acquisition-parameter setting; seed partitions and parameters are listed in the supplement.

In the cached DINO-WM component analyses, every method receives the same random scout queries and spends the remaining budget under its own selection rule. We compare best discovered search score and reference-component coverage on the same top-score set. Sharing the scout queries isolates the continuation policy from variation in the initial observations.

Search baselines are random search, GP-UCB without a typed frontier, and, in cached ablations, the cross-entropy method (CEM) \cite{rubinstein1999cem}. Comparisons match budgets, candidate libraries, and restart streams. Validation uses protocol-specific held-out references; neighborhoods around mid-score and random anchors provide local controls. One-sided exact sign tests assess paired scalar outcomes, McNemar tests paired hit indicators, and exact permutation tests compare prespecified neighborhood groups; ties are omitted.

\begin{figure*}[t]
\centering
\includegraphics[width=.98\textwidth]{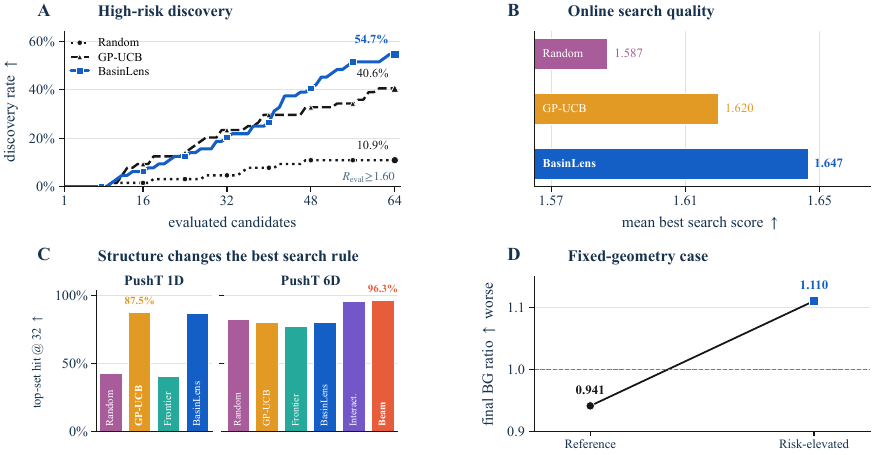}
\caption{Discovery and structure-sensitive search evidence. (A) Cumulative probability of crossing $R_{\mathrm{eval}}\geq1.60$ on the 12{,}016-candidate large online grid. (B) Mean best search score at budget 32 on the 3{,}510-candidate wide validation grid. (C) Cached budget-32 hit rates for the top $2\%$ of the single-factor grid and top $5\%$ of the combinatorial grid: GP-UCB leads in the former, whereas the interaction-aware and beam variants lead in the latter. (D) With block and goal fixed, changing only the valid agent start increases the predicted final block--goal distance ratio; higher is worse.}
\label{fig:search-results}
\end{figure*}

\begin{figure*}[t]
\centering
\includegraphics[width=.98\textwidth]{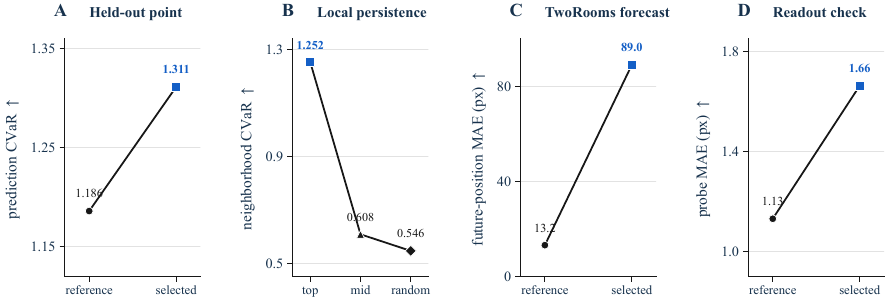}
\caption{Independent validation and measurement boundary. (A) \method{}-selected inputs retain higher mean prediction conditional value at risk (CVaR) on held-out seeds than the default-reset reference. (B) Fresh radius-1 neighborhoods around top-score anchors have higher mean prediction CVaR than neighborhoods around mid-score and random controls, supporting local persistence. (C) In a deterministic TwoRooms case, changing only the valid start raises future-position mean absolute error (MAE) from $13.2$ to $89.0$ pixels. (D) Probe MAE changes only from $1.13$ to $1.66$ pixels in the same case; readout error alone therefore does not explain the forecast error. Panels C and D use different vertical scales.}
\label{fig:validation-boundary}
\end{figure*}

\section{Results}

\subsection{Discovery under a Finite Budget}
On the 12,016-candidate large online grid, random search crosses $R_{\mathrm{eval}}\geq1.60$ in only $10.9\%$ of budget-64 runs, compared with $40.6\%$ for GP-UCB and $54.7\%$ for \method{} (Figure~\ref{fig:search-results}A). Paired McNemar tests give $p=4.18{\times}10^{-6}$ against random search and $p=0.061$ against GP-UCB; the latter fixed-threshold difference is therefore descriptive. This rate indicates that random search rarely reaches the threshold within 64 queries in the finite library. It does not estimate failure prevalence under deployment.

Mean-best and threshold-averaged measures give complementary evidence. At budget 64, mean best search scores are $1.572$, $1.547$, and $1.512$ for \method, GP-UCB, and random search; the paired comparison with GP-UCB gives $p=0.0027$. Averaged over 16 prespecified thresholds from $1.50$ to $1.65$, the corresponding discovery rates are $0.572$, $0.474$, and $0.255$, with $p=0.004$ against GP-UCB. Full budget-wise statistics appear in the supplement.

\subsection{Reproducibility and Local Persistence}
The point-validation protocol evaluates shortlisted inputs on 16 disjoint seeds. The pooled 128-restart analysis contains 27 unique \method{} selections, of which 24 have complete held-out measurements under the prespecified validation caps. Their mean held-out prediction CVaR is $1.311$, compared with $1.186$ for the default reset evaluated on the same seeds (Figure~\ref{fig:validation-boundary}A). Because this protocol does not evaluate other methods' shortlists, the result establishes reproducibility but not comparative shortlist quality.

Fresh-seed evaluation on the wide validation grid tests local persistence. At budget 32, \method{} has the highest mean best search score, and an independent seed split preserves the ordering (Figure~\ref{fig:search-results}B). Radius-1 neighborhoods around the eight top-score anchors have mean prediction CVaR $1.252$, versus $0.608$ around mid-score controls and $0.546$ around random controls (Figure~\ref{fig:validation-boundary}B). Both exact permutation tests satisfy $p\leq6.99{\times}10^{-4}$. The tests treat anchors, rather than individual neighbors, as independent units.

\subsection{What the Discovered Failures Reveal}
Holding the PushT block and goal fixed while changing only the valid agent start raises the predicted final block--goal distance ratio from $0.941$ to $1.110$ across 64 paired seeds; the ratio is higher in $98.4\%$ of pairs (Figure~\ref{fig:search-results}D). This physical readout illustrates the effect of one selected condition. In TwoRooms, changing only the valid start produces a much larger change in future-position error than in probe error (Figure~\ref{fig:validation-boundary}C--D). Across the qualitative cases, failures appear as temporal collapse in TwoRooms, prediction stalling in PushT, and a missed wall-opening transition in DINO-WM. A prioritized local Wall library contains 92 no-cross cases among 900 valid configurations; matched trajectories are visualized in the supplement.

\subsection{When Does Typed Structure Help?}
Matched cached experiments isolate search components at budget 32 (Figure~\ref{fig:search-results}C). On the single-factor grid, GP-UCB attains the highest top-2\% hit rate, $0.875$, followed by \method{} at $0.867$. On the cached combinatorial grid, \beam{} and \interaction{} reach top-5\% hit rates of $0.963$ and $0.961$; random search is the strongest standard baseline at $0.828$, while standard \method{} reaches $0.803$. Thus, pairwise features and multi-anchor expansion help in this combinatorial setting, whereas GP-UCB remains preferable in the single-factor setting.

In DINO-WM common-scout replays, \method{} reaches mean best score $2.908$ versus $2.863$ for GP-UCB and top-5\% reference-component coverage $4.898$ versus $3.195$; an independent 384-candidate library preserves both advantages. \frontier{} attains slightly higher coverage, indicating that typed replacement is the useful component for covering multiple high-risk inputs from one reference component. On JEPA-WM PointMaze, typed action-neighborhood search reaches mean best composite risk $5.113$ at budget 16, versus $4.935$ for random search. In four DIAMOND and IRIS action-prefix studies, random search or UCB is strongest.

Across the studied libraries, typed expansion is most useful when valid inputs contain semantically distinct coordinates and combinatorial interactions. Its advantage is smaller in single-factor and homogeneous action-prefix spaces. Because model, task, and prediction interface vary together, this pattern should be interpreted as a scope observation rather than a controlled causal comparison. The cross-model studies therefore characterize where typed search is useful across prediction interfaces.

\subsection{Scope and Boundaries}
Natural-input failure discovery targets executable prediction failures under environment-valid inputs. Selected PushT conditions are reevaluated on disjoint seeds and in fresh legal neighborhoods. This protocol filters out invalid perturbations and failures caused by a single noisy evaluation. The finite libraries do not estimate failure prevalence under deployment, and risk values are not comparable across prediction interfaces. Search performance also varies with the structure of the input space. 

Each reported failure record should include the environment-valid trigger, search budget, prediction metric, held-out result, and behavior of nearby valid inputs. This record distinguishes a repeatable local weakness from an invalid perturbation or an isolated noisy maximum. The search policy constructs these records, and its effectiveness depends on the input structure. Standard search can remain competitive in low-dimensional or homogeneous spaces, whereas typed replacement is most useful when coordinates have distinct semantics and interactions.

\section{Related Work}
Planning systems use latent-space online planning \cite{hafner2019planet}, latent imagination for policy learning \cite{hafner2020dreamer}, task-relevant learned dynamics \cite{schrittwieser2020muzero}, or latent model predictive control \cite{hansen2022tdmpc}. Their interfaces span discrete-token space \cite{micheli2023iris}, pixel space via diffusion \cite{alonso2024diamond}, and pretrained visual-feature space \cite{zhou2024dinowm}. Benchmarks assess physical fidelity \cite{upadhyay2026worldbench}, interactive behavior \cite{kong2026worldlens,fang2026iworldbench}, and closed-loop outcomes \cite{li2026dworldeval,zhang2026worldinworld}, while uncertainty monitors robot failures \cite{ward2026foundationalwm}. We instead seek environment-valid triggers of model-specific future-prediction risk under fixed models.

\begin{table}[t]
\centering
\footnotesize
\setlength{\tabcolsep}{3pt}
\caption{Positioning relative to adjacent evaluation settings.
Individual methods may combine multiple elements.}
\label{tab:positioning}
\begin{tabular}{@{}p{0.38\linewidth}p{0.56\linewidth}@{}}
\toprule
Paradigm & Typical output \\
\midrule
Aggregate benchmarks &
Aggregate performance over sampled rollouts \\
Adversarial testing &
High-loss cases under optimized perturbations or conditions \\
Policy stress testing &
Policy failures in valid scenarios \\
Scenario falsification &
Counterexamples to stated requirements \\
Black-box optimization &
Best queried inputs for scalar objectives \\
This work &
Environment-valid triggers with held-out point and
legal-neighborhood evidence \\
\bottomrule
\end{tabular}
\end{table}
Software testing uses combinatorial designs \cite{kuhn2013combinatorial}, while neural-system testing uses coverage \cite{pei2017deepxplore}, metamorphic transformations \cite{tian2018deeptest}, and behavioral templates \cite{ribeiro2020checklist}. Rare-failure evaluation and adaptive stress testing target agent or policy failures \cite{uesato2019rigorous,koren2018ast}. VerifAI searches Scenic-defined scenarios for requirement violations \cite{dreossi2019verifai,fremont2019scenic}, and related falsification methods support robust reinforcement learning \cite{wang2020frarl}. Bayesian safety validation estimates failure probability \cite{moss2024bayesiansafety}, whereas Bayesian-optimization falsification targets requirement violations \cite{ramezani2025bofalsification}. Natural-input failure discovery searches a fixed world model's predictions and attaches independent point and neighborhood evidence to returned conditions.

Adversarial examples perturb inputs \cite{goodfellow2015adversarial}, physical attacks alter sensed objects \cite{eykholt2018physical}, and adversarial reinforcement learning attacks policies through perturbations or opponent actions \cite{huang2017adversarialrl,gleave2020adversarialpolicies}. World-model attacks perturb physical-conditioning channels or search attack configurations \cite{guo2026physcondwma,guo2026wmattack}; the Tail-Aware Ranking Attack for World-Model Planning (TRAP) targets imagined-trajectory rankings \cite{duan2026trap}. GP-UCB searches expensive black-box objectives \cite{srinivas2010gpucb}. We instead search for valid inputs that expose prediction failures and validate those failures across seeds and nearby legal edits. \method{} combines GP-UCB coverage with typed replacements for this purpose.

\section{Conclusion}
We introduced natural-input failure discovery, a finite-budget
evaluation problem for locating environment-valid resets or action
prefixes that induce reproducible, locally persistent prediction
failures. Across PushT and four additional world-model settings, the
results suggest that typed expansion is most useful for semantically
distinct, interacting coordinates, while standard search remains
competitive in simpler spaces. \method{} provides a structured
implementation of this problem and complements aggregate benchmarks
by turning hidden weaknesses into inspectable failure records supported
by held-out and neighborhood evidence.

\clearpage
\bibliography{references,references_additions}

\clearpage
\appendix
\section*{Appendix}
\numberwithin{figure}{section}
\numberwithin{table}{section}
\numberwithin{equation}{section}
\section{Supplement Overview}
The experiments answer complementary questions with separate protocols. The 12,016-candidate large online grid measures discovery; a 576-candidate library reevaluates shortlisted points; two splits of the 3,510-candidate wide validation grid measure search replication and fresh-neighborhood risk; cached libraries isolate search components; and seed-matched planning compares controller outcomes under different ranking rules. The first three stages form the main evidence chain. Component and cross-model studies test method behavior and interface scope, while planning is an independent consequence diagnostic. All differences and gains are computed from unrounded run-level values; displayed summaries are rounded.

\begin{table*}[!t]
\centering
\footnotesize
\setlength{\tabcolsep}{4pt}
\begin{tabularx}{\textwidth}{@{}p{1.08in}p{.78in}p{1.12in}p{.70in}X@{}}
\toprule
Question & Model/task & Valid-input library & Query setting & Evidence returned \\
\midrule
Discovery & LeWM PushT & Large online grid (12,016) & $B=16,32,64$; 64 restarts & Best search score and threshold crossing under matched budgets \\
Point validation & LeWM PushT & Point-validation library (576) & $B=32$; pooled 128 restarts & Shortlisted points reevaluated on 16 disjoint seeds \\
Local persistence & LeWM PushT & Wide validation grid (3,510) & $B=32$; two seed splits & Fresh radius-1 neighborhoods around top, mid-score, and random anchors \\
Component isolation & LeWM PushT & Cached combinatorial grid (431) & $B=8,16,32$; 512 restarts & Top-set hit rate for global, frontier, interaction, and beam variants \\
Replication & DINO-WM PushT & 512 and 384 geometries & $B=32$; 128 paired replays & Point discovery and reference-component coverage after a common scout \\
Interface scope & JEPA-WM/Atari & 480/128/36 valid inputs & Protocol-specific budgets & Latent-feature or future-frame prediction-risk discovery \\
Consequence & LeWM PushT & Four selected contexts & 8 context seeds & Environment outcomes under model-ranked and oracle-ranked control \\
\bottomrule
\end{tabularx}
\caption{Protocol map. Candidate libraries, seed partitions, and measurements are intentionally distinct because each row answers a different question. Core claims do not combine unmatched protocols into a single statistic.}
\label{tab:supp-protocol-map}
\end{table*}

\section{Reading the Discovery Workflow}
\begin{figure*}[t]
\centering
\includegraphics[width=.99\textwidth]{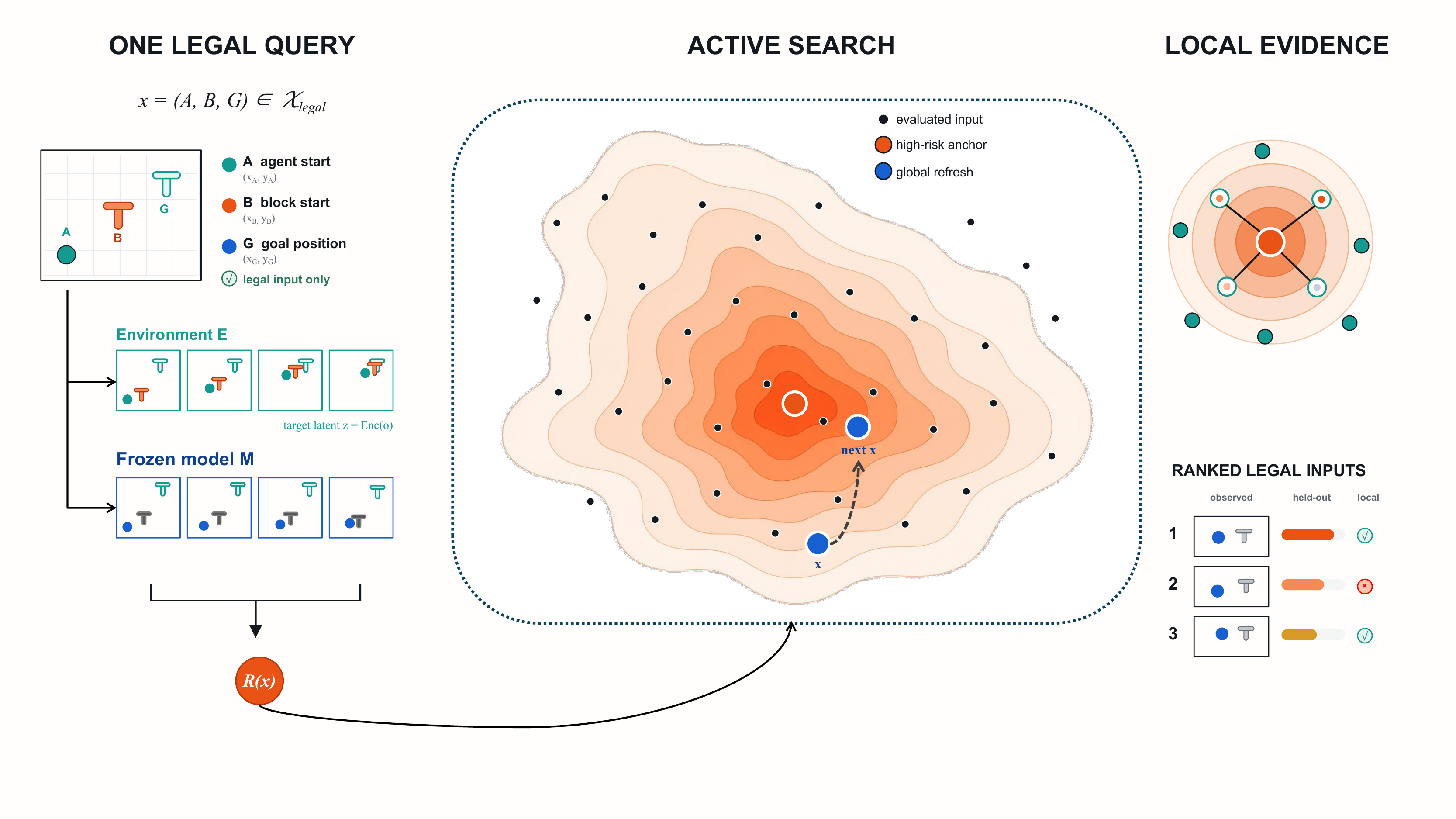}
\caption{Detailed natural-input failure-discovery workflow. Left: one legal typed input is executed in the environment and the frozen world model, producing a prediction-risk observation. Center: active search alternates global refreshes with typed moves around evaluated high-risk inputs while querying only members of the legal input set. Right: returned inputs are ranked and evaluated using held-out seeds and nearby legal changes. The orange contours are a schematic risk landscape, not an oracle surface available to the search algorithm.}
\label{fig:supp-detailed-workflow}
\end{figure*}

Figure~\ref{fig:supp-detailed-workflow} expands the compact overview in the main paper. It separates the object being tested, the budgeted search process, and the evidence returned after discovery. This separation is essential: a valid query defines an executable test condition; an acquisition rule determines which conditions are evaluated; and held-out or local evaluation determines what can be concluded about the returned condition.

The schematic PushT input $x=(A,B,G)$ contains an agent start, a block start, and a goal position. These coordinates are ordinary reset variables accepted by the environment; the search neither perturbs pixels nor constructs an invalid simulator state. Other model interfaces replace this tuple with their own typed specification, such as a reset seed and action template. In every case, the environment execution supplies the matched future used to evaluate the fixed world model.

The shaded center region visualizes the motivation for local expansion: nearby valid inputs may share elevated prediction risk. The algorithm never observes this complete landscape. It sees only queried risks, uses a global surrogate proposal to avoid premature localization, and expands around observed anchors through valid one-coordinate replacements. The ranked output is therefore a finite discovery record, not a claim that all high-risk regions have been enumerated. Held-out reevaluation and fresh-neighborhood evaluation are run after search and do not change the recorded query sequence.

\section{\method{} Implementation}
Each coordinate is min--max normalized over the valid candidate library. We denote by $\mathcal{R}(x)$ the search frontier obtained by replacing one typed coordinate with any other value available in that coordinate's library and retaining the candidate only if the resulting tuple belongs to the valid library. Fresh radius-1 neighborhood evaluation is narrower: it changes one coordinate by one adjacent grid level. Thus, position, goal, and action-prefix coordinates are never assigned values outside their prespecified input specification, while search expansion and validation locality remain distinct. Stable candidate identifiers are assigned before normalization, so the discrete input and every permitted replacement remain reconstructible independently of the surrogate representation.

For the merged-pool score, let $\overline{R}_t$ and $s_t^R=\max(\operatorname{std}\{R_{\mathrm{eval}}(x):x\in\mathcal{Q}_t\},10^{-6})$ be the mean and standard deviation of the observed search scores. The upper confidence bound (UCB) and structural terms from the main paper are
\begin{align}
z^{\mathrm{UCB}}_t(x)
  &=\frac{\mu_t(x)+\beta\sigma_t(x)-\overline{R}_t}{s_t^R},\\
z^{\mathrm{par}}_t(x)
  &=\max_{h\in\mathcal{H}_t:\,x\in\mathcal{R}(h)}
    \frac{R_{\mathrm{eval}}(h)-\overline{R}_t}{s_t^R},\\
\kappa^{\mathrm{anc}}_t(x)
  &=\max_{h\in\mathcal{H}_t}
    \exp\!\left(-\frac{\|\psi(x)-\psi(h)\|_2^2}{2s^2}\right),\\
\delta^{\mathrm{nov}}_t(x)
  &=\min_{q\in\mathcal{Q}_t}\|\psi(x)-\psi(q)\|_2 .
\end{align}
For a global candidate with no frontier parent, the implementation sets $z^{\mathrm{par}}_t(x)=-3$. Ties in the complete score are broken by the algorithm-restart random-number generator.

The LeWorldModel (LeWM) and DINO World Model (DINO-WM) implementations use a Gaussian process (GP) with a constant kernel fixed at 1 multiplied by a radial-basis-function kernel, plus a fixed white-noise kernel of $10^{-4}$. Each normalized coordinate has initial length scale $0.25$ with bounds $[0.03,3.0]$; the regressor uses $\alpha=10^{-6}$, normalized targets, and no optimizer restarts. Table~\ref{tab:supp-implementation} lists the remaining parameters. The LeWM column is used unchanged for the 576-, 3,510-, and 12,016-candidate online PushT protocols; DINO-WM uses the separate setting shown in its column. Here, $b_0$ is the scout size, Anchors is $|\mathcal{H}_t|$, Global pool is $|\mathcal{P}^{\mathrm{glob}}_t|$, and $s$ is the anchor-proximity scale.

\begin{table}[!htbp]
\centering
\small
\setlength{\tabcolsep}{1.5pt}
\begin{tabularx}{\columnwidth}{@{}Xrr@{}}
\toprule
Parameter & LeWM PushT & DINO-WM PushT \\
\midrule
Scout size $b_0$ & 3 & 8 \\
UCB exploration coefficient $\beta$ & 0.35 & 0.45 \\
Number of anchors & 4 & 4 \\
Global-pool size & 16 & 24 \\
Parent weight $\lambda_p$ & 0.10 & 0.10 \\
Proximity weight $\lambda_d$ & 0.08 & 0.12 \\
Novelty weight $\lambda_n$ & 0.05 & 0.05 \\
Proximity scale $s$ & 0.50 & 0.45 \\
\bottomrule
\end{tabularx}

\caption{Parameters used by the LeWM online protocols and the DINO-WM common-scout replay.}
\label{tab:supp-implementation}
\end{table}

\interaction{} fits a ridge predictor with penalty $1.0$ to normalized linear, squared, and pairwise-product features in the cached combinatorial analysis. At each round, it scores every unseen input by its predicted search score plus $0.30s_R$ times distance to the nearest observation and $0.25s_R$ times Gaussian proximity to observed anchors, where $s_R$ is the robust observed-score scale, the proximity scale is $0.18$, and the anchors are observations at or above the current upper quartile (the maximum before four observations). \beam{} is a separate frontier-only diagnostic with width four. At each query, it recomputes the four highest-scoring observed anchors, collects every valid one-coordinate replacement, and omits the global candidate pool. After eight observations, a ridge predictor with penalty $2.0$ scores linear, squared, and pairwise coordinate features; before then, its predicted-score term is zero. Its acquisition is $0.55$ times standardized parent score plus $0.30$ times standardized predicted score, $0.30$ times novelty, and $0.25$ times anchor proximity. It evaluates one maximizer and then recomputes the anchors. These variants are reported as component-isolation tools; the online 12,016- and 3,510-candidate results use \method{}.

The cached component analysis initializes the cross-entropy method (CEM) with three random inputs. At each subsequent query, the best $\lceil0.35t\rceil$ of the $t$ observations are elites. For every unseen normalized input, CEM computes the minimum Euclidean distance $d$ to an elite, assigns unnormalized weight $0.85\exp[-d^2/(2\cdot0.22^2)]+0.15$, and samples one input from the normalized weights.

\subsection{Complete Discovery Procedure}
The implementation follows the same execution order in every finite-library protocol; only the model-specific evaluator, input specification, and parameters change.

\begin{enumerate}
\item \textbf{Construct and verify the library.} Enumerate all tuples from the typed input specification, apply the environment-specific validity filters, and assign each retained tuple a stable candidate identifier. Candidate values, identifiers, and normalized coordinates are fixed before any search restart.
\item \textbf{Create a paired scout.} For algorithm-restart seed $s$, sample the first $b_0$ candidate identifiers without replacement. Every method in a matched comparison receives the same scout order and the same stored evaluation for each scout candidate. Thus, differences after the scout arise from the continuation policy rather than initial observations.
\item \textbf{Evaluate a query.} Execute the candidate on the prespecified seed set, compute each per-seed prediction loss, aggregate it into prediction risk, and apply the task-activity adjustment when required. The resulting record stores the typed input, per-seed losses, prediction risk, activity measurements, and search score. Previously evaluated candidate identifiers are never queried again within a restart.
\item \textbf{Build global and local pools.} Fit the surrogate to all records available in the current restart and select the highest-UCB unseen candidates for the global pool. Separately, take the current high-score anchors, enumerate their valid one-coordinate replacements, and remove all evaluated identifiers to obtain the local pool. Pool membership is recomputed after each query.
\item \textbf{Rank the merged pool.} Compute the acquisition $A_t(x)$ for every candidate in the union. The highest acquisition value determines the next query. Exact acquisition ties are resolved using the restart-specific random-number generator, which avoids dependence on incidental file order while preserving complete replayability from the seed.
\item \textbf{Return a discovery record.} After $B$ queries, rank evaluated candidates by search score using a stable identifier-based tie rule. The record contains the entire query sequence and the top-$k$ shortlist; it does not silently replace the final choice with an oracle scan of the full library.
\item \textbf{Run follow-up evaluation separately.} Held-out point and neighborhood measurements use their own seed sets and do not update the completed search trace. This separation prevents validation outcomes from leaking back into acquisition or shortlist ranking.
\end{enumerate}

For cached replay, Step 3 reads a fixed candidate record rather than executing a new model--environment rollout. All compared methods read the same cache entry for a given candidate, so the replay changes only the order in which candidates are revealed. For online protocols, candidate evaluations are written once and reused by all post-processing summaries. Summary tables are computed from unrounded run-level values.

\section{Protocol Details}
The main protocols use LeWorldModel (LeWM) \cite{maes2026lewm} on PushT \cite{chi2023diffusionpolicy,florence2022implicit}. Cross-interface analyses use DINO World Model (DINO-WM) \cite{zhou2024dinowm}, a joint-embedding predictive architecture world model (JEPA-WM) \cite{terver2025jepawms}, DIAMOND \cite{alonso2024diamond}, and IRIS \cite{micheli2023iris}. The shared formulation uses each interface's prediction target: state-probe output for LeWM, visual features for DINO-WM, joint embeddings for JEPA-WM, directly generated frames for DIAMOND, and decoded RGB frames from discrete tokens for IRIS.

\begin{table*}[t]
\centering
\small
\setlength{\tabcolsep}{4pt}
\begin{tabularx}{\textwidth}{@{}p{.88in}p{1.38in}p{1.55in}X@{}}
\toprule
Model/task & Typed natural input & Validity rule & Prediction-risk interface \\
\midrule
LeWM PushT & Agent, block, and goal $x/y$ coordinates & Candidate must instantiate a simulator reset and satisfy the protocol's minimum agent--block and block--goal distances & Upper-tail CVaR of final-step state-probe MSE \\
LeWM TwoRooms & Agent start $x/y$ with fixed target & Start lies in navigable free space under the fixed room geometry & Predicted-future position MAE and target-distance gap \\
DINO-WM PushT & Agent, block, and goal $x/y$ coordinates & Simulator-valid reset with minimum geometry distances & Upper-tail CVaR of final-step visual-feature MSE \\
DINO-WM Wall & Start, goal, and opening geometry & Simulator-valid local configuration with a matched environment rollout & Decoded-path discrepancy and visible crossing outcome \\
JEPA-WM PointMaze & Reset seed, action template, and scale & Scaled actions remain in $[-1,1]^2$ and use the fixed action horizon & Visual-latent MSE plus $0.1$ proprioceptive-latent MSE \\
DIAMOND Atari & Reset seed and four discrete prefix actions & Every prefix entry belongs to the game's action set & One-step future-frame MSE \\
IRIS Atari & Reset seed, no-op count, and action template & Nonnegative no-op count and game-valid template actions & Four-step future-frame MSE \\
\bottomrule
\end{tabularx}
\caption{Typed input specifications and model-specific prediction risks. A local replacement changes one listed input field and retains the candidate only if the corresponding validity rule still holds.}
\label{tab:supp-input-interfaces}
\end{table*}

The experiments were run on Debian GNU/Linux 12 with two Intel Xeon Platinum 8558 processors (192 logical CPUs), 2.0 TiB of system memory, and an NVIDIA H20-3e GPU with 143771 MiB of memory and driver 575.57.08. The recorded environment used uv 0.10.9 and Python 3.12.13. Key package versions were PyTorch 2.9.1, NumPy 2.4.1, SciPy 1.17.0, scikit-learn 1.8.0, Matplotlib 3.10.8, Transformers 4.57.3, Gymnasium 1.2.3, MuJoCo 3.6.0, Pymunk 7.2.0, TensorDict 0.11.0, Einops 0.8.2, and Pillow 12.1.0. The LeWM implementation uses stable-worldmodel 0.0.6 pinned to commit prefix \texttt{ba10600b31ab}. The code supplement provides \texttt{pyproject.toml}, \texttt{uv.lock}, and the complete recorded package list.

The anonymous code supplement contains the core discovery, validation, neighborhood, component-isolation, probe, controller, and cross-model entry points, together with their local dependencies and seed specifications. Model-specific adapters require the cited upstream repositories and pretrained checkpoints; these external assets are not duplicated because their distribution and licenses remain controlled by their original maintainers. The adapters accept explicit repository and checkpoint paths.

Validity is an environment-level property; locality is defined only after invalid tuples have been removed. The search frontier may replace one coordinate with any allowed value, which enables larger moves within the same typed factor. Radius-1 validation is stricter and uses only an adjacent value in the coordinate's ordered grid. Consequently, every radius-1 neighbor is valid by construction, but not every valid one-coordinate replacement is a radius-1 neighbor. For categorical action templates without an intrinsic ordering, the protocol reports typed replacement results rather than claiming geometric radius.

Table~\ref{tab:supp-seed-ledger} records every seed partition used in the principal LeWM PushT evidence chain. Search, point validation, and neighborhood evaluation use disjoint environment-seed sets within each protocol. Algorithm-restart seeds determine scout order and acquisition tie breaking; they do not replace environment seeds.

\begin{table*}[t]
\centering
\small
\setlength{\tabcolsep}{4pt}
\begin{tabularx}{\textwidth}{@{}p{1.15in}p{.82in}p{.82in}p{.92in}p{.92in}X@{}}
\toprule
Protocol & Restart seeds & Search-evaluation seeds & Point-validation seeds & Neighborhood seeds & Notes \\
\midrule
576-candidate search & 32--127 & 0--7 & -- & -- & Scout size three; budget 32 \\
576-candidate pooled validation & 0--127 & 0--7 & 16--31 & -- & 96 runs use scout size three and 32 use scout size four; 24 of 27 unique \method{} selections have complete held-out measurements \\
3,510-candidate split A & 192--255 & 0--7 & 32--47 & 48--63 & Budget 32; point and radius-1 neighborhood records are kept separate \\
3,510-candidate split B & 256--383 & 16--23 & 80--95 & 96--111 & Independent restart and environment-seed split; budget 32 \\
12,016-candidate discovery & 384--447 & 24--31 & -- & -- & Budgets 16, 32, and 64 \\
\bottomrule
\end{tabularx}
\caption{Seed ledger for the principal LeWM PushT protocols. A dash indicates that the corresponding follow-up evaluation is not part of that protocol.}
\label{tab:supp-seed-ledger}
\end{table*}

The 576-candidate reference is the simulator reset obtained without a geometry override and reevaluated on the same held-out seeds as the selected candidate. Each pooled-validation run has a prespecified validation cap; incomplete availability is reported rather than filled by reusing search seeds.

The 576-candidate protocol uses agent and goal positions $\{96,256,416\}$ and block positions $\{128,256,384\}$ on each axis. It retains Cartesian combinations whose agent--block and block--goal center distances are both at least 80 pixels.

The 12,016-candidate library is generated from six typed coordinates. Agent $x/y$ positions use $\{56,116,176,236,296\}$; block $x/y$ positions use $\{112,176,240,304,368\}$; and goal $x/y$ positions use $\{80,160,240,320,400\}$. From the $5^6=15,625$ Cartesian combinations, the input specification removes inputs whose agent--block or block--goal center distance is below 80 pixels, leaving 12,016 valid candidates.

The 3,510-candidate library uses the same six coordinate types and validity filter. Agent positions use $\{56,176,296,416\}$, block positions use $\{112,208,304,400\}$, and goal positions use $\{96,200,304,416\}$ on each axis. Filtering the $4^6=4,096$ Cartesian combinations by the same 80-pixel agent--block and block--goal distance constraints leaves 3,510 valid candidates.

The 576-, 3,510-, and 12,016-candidate online protocols use final-step conditional value at risk (CVaR) as the raw prediction risk $R_{\mathrm{pred}}$. Let $b_{\mathrm{prog}}$, $b_{\mathrm{con}}$, and $b_{\mathrm{path}}$ be the default reset's mean block--goal progress ratio, total contacts, and block path length on the same evaluation seeds. The task-activity floors are
\[
\begin{aligned}
f_{\mathrm{prog}}&=\max\{-1,b_{\mathrm{prog}}-0.15\},\\
f_{\mathrm{con}}&=0.5b_{\mathrm{con}},\qquad
f_{\mathrm{path}}=0.5b_{\mathrm{path}}.
\end{aligned}
\]
For candidate measurements $r_{\mathrm{prog}}$, $n_{\mathrm{con}}$, and $L_{\mathrm{path}}$ in the same order, the normalized violations are
\[
\begin{aligned}
v_{\mathrm{prog}}&=\frac{[f_{\mathrm{prog}}-r_{\mathrm{prog}}]_+}
{\max(0.1,|f_{\mathrm{prog}}|)},\\
v_{\mathrm{con}}&=\frac{[f_{\mathrm{con}}-n_{\mathrm{con}}]_+}
{\max(1,f_{\mathrm{con}})},\\
v_{\mathrm{path}}&=\frac{[f_{\mathrm{path}}-L_{\mathrm{path}}]_+}
{\max(1,f_{\mathrm{path}})}.
\end{aligned}
\]
where $[z]_+=\max(z,0)$. Online acquisition uses $R_{\mathrm{eval}}=R_{\mathrm{pred}}-20(v_{\mathrm{prog}}+v_{\mathrm{con}}+v_{\mathrm{path}})$. A candidate is task-active when all three violations are zero. The reported best is selected among task-active evaluated candidates whenever that set is nonempty and otherwise among all evaluated candidates. Every reported best is task-active in these analyses, so the displayed best scores equal unpenalized CVaR; held-out point and neighborhood tables also report unpenalized CVaR.

\begin{table*}[t]
\centering
\small
\setlength{\tabcolsep}{5pt}
\begin{tabularx}{\textwidth}{@{}p{1.02in}p{1.12in}X X@{}}
\toprule
Term & Symbol or endpoint & Definition & Permitted interpretation \\
\midrule
Prediction risk & $R_{\mathrm{pred}}$ or $R_{\mathrm{val}}$ & Aggregate discrepancy between world-model predictions and matched environment futures & Prediction quality under the specified model interface and seed set \\
Search score & $R_{\mathrm{eval}}$ & Prediction risk minus a prespecified task-activity penalty & Acquisition and shortlist ranking; equal to prediction risk only for task-active candidates \\
Task consequence & Environment progress & Closed-loop block--goal progress under a controller & Outcome of the controller protocol; not another prediction-risk measurement \\
\bottomrule
\end{tabularx}
\caption{Terminology used throughout the paper. Prediction risk, search score, and task consequence are distinct quantities and are not substituted for one another.}
\label{tab:supp-metric-separation}
\end{table*}

For each \method{} budget-32 restart, the protocol records the best selected candidate; repeated candidate identifiers are merged by retaining their largest search score, and the eight highest-scoring unique candidates form the reported top-score anchor group. A Gaussian-process upper confidence bound (GP-UCB) top-eight set is constructed identically but is used only when constructing exclusions. The mid-score controls are the eight central entries after restricting the source search's online evaluations to records marked task-active, deduplicating by candidate identifier with the final recorded search score, and sorting by that score. Stable source order breaks equal scores. A random-number generator with seed 20260601 samples eight random controls without replacement from the full valid library after excluding the \method{} and GP-UCB top sets and the mid-score controls. Each anchor statistic is the mean candidate-level conditional value at risk (CVaR) over the anchor and its unique valid radius-1 neighbors; a candidate shared by different anchors or groups remains in each corresponding anchor statistic. The one-sided exact permutation test treats the eight anchor statistics in each group as its units and enumerates all $\binom{16}{8}$ reallocations.

The PushT fixed-geometry case holds the block at $(128,256)$ and goal at $(256,256)$ while changing the agent start from $(56,96)$ to $(256,256)$. The final block--goal (BG) ratio is the final predicted block--goal center distance divided by its initial distance, so a larger value indicates less predicted task progress. The ratio is a model-predicted task-outcome readout rather than $R_{\mathrm{pred}}$ or $R_{\mathrm{eval}}$. The two starts are compared on paired rollout seeds. In deterministic TwoRooms, the target is fixed at $(168,196)$ while the agent start varies; reference and risk-elevated starts have predicted-future $x$--$y$ position mean absolute errors (MAEs) of $13.2$ and $89.0$ pixels (px) and final target-distance gaps of $27.2$ and $149.3$ px. Probe MAEs on environment-encoded latents are $1.13$ and $1.66$ px.

DINO-WM observes one frame and autoregressively predicts eight visual-feature states sampled at five-simulator-step intervals from a 40-step PushT rollout. For each seed, the raw risk is the final-step mean squared error between the predicted visual features and DINO-encoded simulator-frame features. The 512-candidate library uses agent coordinates $\{56,136,216,296\}$, block coordinates $\{112,208,304,400\}$, and goal coordinates $\{96,192,288,416\}$ on each axis; seed 20260602 controls its random subsampling. The independent 384-candidate library uses $\{56,176,296\}$, $\{112,256,384\}$, and $\{96,256,416\}$, respectively, with subsampling seed 3. Both libraries remove combinations whose agent--block or block--goal distance is below 70 pixels and then sample without replacement to reach the stated size. The 512-candidate replay aggregates evaluation seeds 16--23 with upper-tail CVaR at tail fraction $0.4$, i.e., the mean of the four largest values among eight seeds. The 384-candidate replay uses evaluation seeds 4--7 and the same tail fraction, i.e., the mean of the two largest values among four seeds. A candidate is task-active when its mean contact steps are at least 1, mean block path length is at least 10 pixels, and mean block--goal progress ratio is at least $-1$. Its stored replay score $R_{\mathrm{eval}}$ equals CVaR when task-active and CVaR minus 2 otherwise. Common-scout acquisition, best-score reporting, and oracle top-score sets all use this stored score. Let $\mathcal{G}_{\mathcal{R}}$ be the valid-input graph that connects inputs differing by one arbitrary allowed coordinate replacement. For an evaluated trace $\mathcal{T}$ and oracle top-$q\%$ score set $\mathcal{H}_q$, reference-component coverage is $C_q(\mathcal{T})=\max_{C\in\operatorname{Comp}_{\mathcal{G}_{\mathcal{R}}}(\mathcal{H}_q)}|\mathcal{T}\cap C|$, or zero if no top-score input is evaluated. Components are computed on the full reference subgraph, not on the trace-induced subgraph.

The LeWM checkpoints lack a red--green--blue (RGB) decoder, so a latent-to-state probe supplies both the reported PushT state mean squared error (MSE) risk and interpretable state readouts. Table~\ref{tab:supp-probe} quantifies the probe's held-out, in-distribution accuracy. The LeWM results therefore evaluate the complete world-model-to-state-readout pipeline.

\begin{table}[!htbp]
\centering
\small
\setlength{\tabcolsep}{2pt}
\begin{tabular}{llllrr}
\toprule
Model & Task & Target & Unit & MAE $\downarrow$ & $N$ \\
\midrule
LeWM & PushT & Agent position & px & 19.64 & 48 \\
LeWM & PushT & Block position & px & 8.90 & 48 \\
LeWM & PushT & Block angle & deg & 3.84 & 48 \\
LeWM & TwoRooms & Agent position & px & 1.55 & 11 \\
\bottomrule
\end{tabular}

\caption{Held-out probe checks for LeWM tasks without native RGB decoding. MAE denotes mean absolute error, px denotes pixels, deg denotes degrees, and $N$ is the number of held-out samples; lower is better.}
\label{tab:supp-probe}
\end{table}

\subsection{Probe-Specific Diagnostic and Scope}
\begin{figure*}[t]
\centering
\includegraphics[width=.98\textwidth]{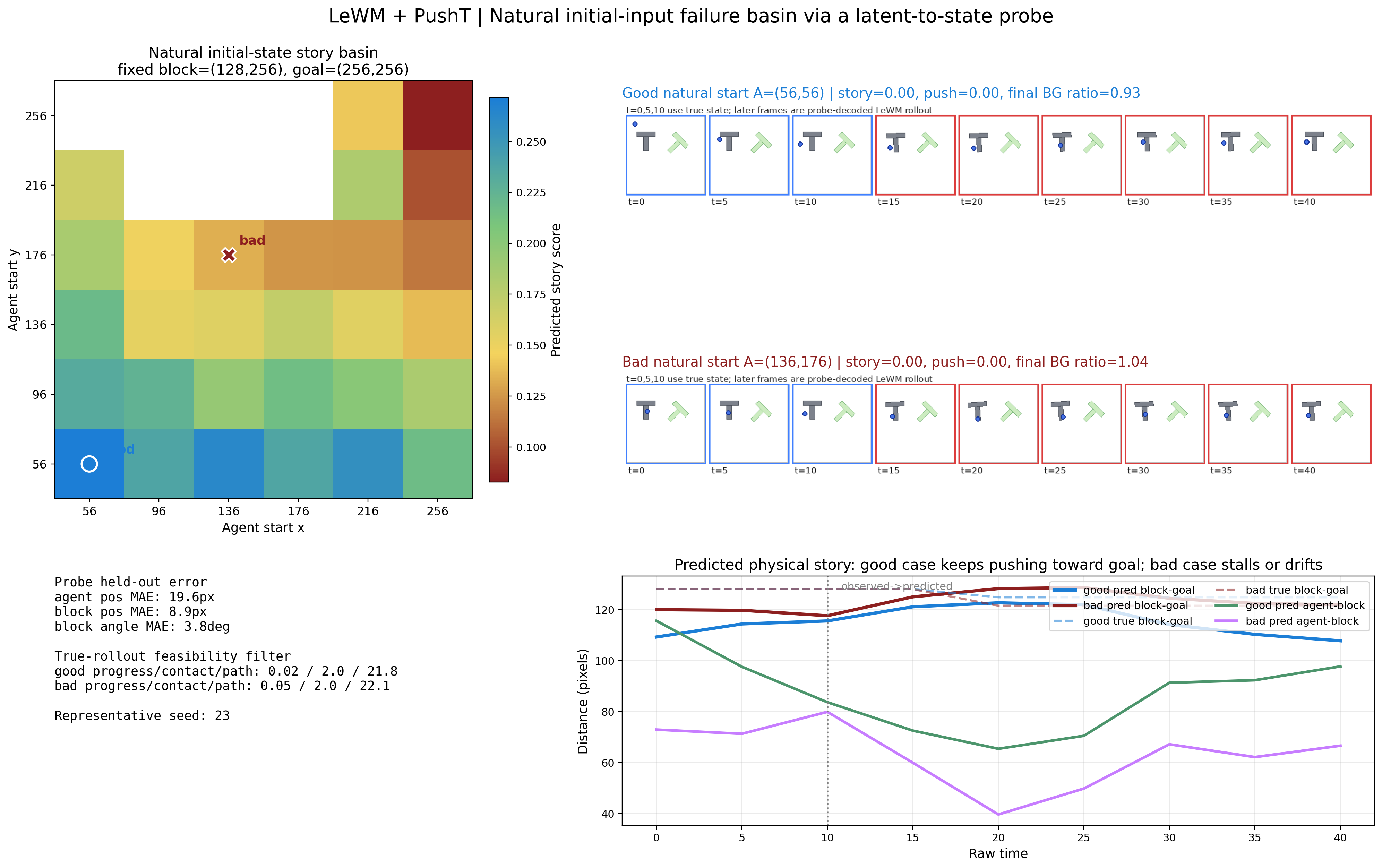}
\caption{LeWM PushT probe diagnostic under fixed block--goal geometry. The left panel shows the historical legal-start scan; the upper and lower strips compare a lower-score and a higher-score start for one representative seed; and the lower-right plot shows physical distances decoded from the predicted latent sequence. The listed held-out probe errors are measured on environment-encoded latents. This figure is illustrative and is not counted as an additional statistical sample.}
\label{fig:supp-probe-diagnostic}
\end{figure*}

Figure~\ref{fig:supp-probe-diagnostic} shows an existing fixed-geometry diagnostic that was generated before the large-grid protocols. It keeps the block and goal fixed, scans valid agent starts, and compares one lower-score start with one higher-score start under the same readout pipeline. The panel is useful because it exposes all layers of the measurement at once: the legal reset, probe-decoded future states, task-activity measurements from the matched environment rollout, and physical distance traces derived from the decoded prediction.

The held-out check establishes that the probe recovers ordinary environment states with the errors listed in Table~\ref{tab:supp-probe}; it does not prove that the same probe is uniformly accurate on every model-predicted latent encountered by search. Consequently, the LeWM result is stated as failure of the complete frozen world-model-to-state-readout pipeline. We do not attribute the entire measured error uniquely to latent dynamics, and the LeWM evidence is complemented by DINO-WM and JEPA-WM protocols that define risk directly in their visual-feature spaces. The fresh-seed and fresh-neighborhood results address repeatability of the measured pipeline failure, not an untestable decomposition of that failure into model and probe components.

\begin{table*}[t]
\centering
\small
\setlength{\tabcolsep}{5pt}
\begin{tabular}{lrrrrrr}
\toprule
Method & Selected unique & Validated & Mean CVaR & Task-active & Contacts & Block path \\
\midrule
Random & 46 & 39 & 1.292 & 1.000 & 10.05 & 93.18 \\
GP-UCB & 34 & 30 & 1.278 & 0.900 & 10.61 & 88.06 \\
\method{} & 27 & 24 & \textbf{1.311} & 0.958 & 11.58 & 94.59 \\
\bottomrule
\end{tabular}

\caption{Method-specific held-out summary for the 576-candidate protocol. Selected unique counts are deduplicated candidates returned under the per-run validation caps; Validated counts have complete measurements on the 16 disjoint seeds. Contacts and block path are validation-seed averages, with block path measured in pixels. Because methods produce different candidate sets and validation availability, this table is descriptive rather than a paired search-performance comparison.}
\label{tab:supp-heldout-by-method}
\end{table*}

\section{Online Discovery and Replication}
\begin{table}[!htbp]
\centering
\small
\setlength{\tabcolsep}{3pt}
\begin{tabular}{llrcc}
\toprule
$B$ & Method & Mean $\pm$ SEM & $p_R$ & $p_G$ \\
\midrule
16 & Random & $1.368{\pm}0.017$ & -- & -- \\
   & GP-UCB & $1.370{\pm}0.027$ & -- & -- \\
   & \method{} & $\mathbf{1.411{\pm}0.023}$ & 0.0056 & 0.295 \\
\midrule
32 & Random & $1.444{\pm}0.014$ & -- & -- \\
   & GP-UCB & $1.481{\pm}0.020$ & -- & -- \\
   & \method{} & $\mathbf{1.493{\pm}0.018}$ & 0.0011 & 0.175 \\
\midrule
64 & Random & $1.512{\pm}0.010$ & -- & -- \\
   & GP-UCB & $1.547{\pm}0.013$ & -- & -- \\
   & \method{} & $\mathbf{1.572{\pm}0.011}$ & $5.56{\times}10^{-6}$ & 0.0027 \\
\bottomrule
\end{tabular}

\caption{Mean best search score on the 12,016-candidate PushT grid across 64 paired restarts (mean $\pm$ standard error of the mean (SEM); higher is better). $p_R$ and $p_G$ compare \method{} with random search and Gaussian-process upper confidence bound (GP-UCB) search.}
\label{tab:supp-largegrid-best}
\end{table}

\begin{table}[!htbp]
\centering
\small
\setlength{\tabcolsep}{3pt}
\begin{tabular}{llrrr}
\toprule
$B$ & Method & Hit rate & \method{} gain & Paired $p$ \\
\midrule
16 & Random & 0.056 & 0.117 & $7.2\times10^{-4}$ \\
   & GP-UCB & 0.148 & 0.024 & 0.229 \\
   & \method{} & \textbf{0.173} & -- & -- \\
\midrule
32 & Random & 0.135 & 0.188 & $7.0\times10^{-4}$ \\
   & GP-UCB & 0.298 & 0.025 & 0.140 \\
   & \method{} & \textbf{0.323} & -- & -- \\
\midrule
64 & Random & 0.255 & 0.317 & $8.7\times10^{-6}$ \\
   & GP-UCB & 0.474 & 0.099 & 0.004 \\
   & \method{} & \textbf{0.572} & -- & -- \\
\bottomrule
\end{tabular}

\caption{Threshold-averaged hit rate on the 12,016-candidate large online PushT grid (higher is better). $B$ is the query budget. Each score averages 16 hit indicators over $R_{\mathrm{eval}}\in[1.50,1.65]$ in increments of 0.01. On a comparator row, \method{} gain is \method{} minus that method, and paired $p$ is the corresponding one-sided exact paired sign-test value.}
\label{tab:supp-largegrid-hit}
\end{table}

\begin{table*}[t]
\centering
\small
\setlength{\tabcolsep}{4pt}
\begin{tabular}{llrrrrrr}
\toprule
$B$ & Threshold & Random & GP-UCB & BasinLens & Gain$_R$ & $p_R$ & $p_G$ \\
\midrule
16 & $R\geq 1.55$ & 0.078 & 0.172 & \textbf{0.219} & 2.8$\times$ & 0.011 & 0.274 \\
16 & $R\geq 1.60$ & 0.016 & \textbf{0.094} & 0.062 & 4.0$\times$ & 0.188 & 0.855 \\
16 & $R\geq 1.62$ & 0.016 & 0.031 & \textbf{0.047} & 3.0$\times$ & 0.312 & 0.500 \\
\midrule
32 & $R\geq 1.55$ & 0.188 & 0.422 & \textbf{0.438} & 2.3$\times$ & 0.003 & 0.500 \\
32 & $R\geq 1.60$ & 0.047 & \textbf{0.234} & 0.203 & 4.3$\times$ & 0.011 & 0.760 \\
32 & $R\geq 1.62$ & 0.031 & 0.078 & \textbf{0.109} & 3.5$\times$ & 0.090 & 0.377 \\
\midrule
64 & $R\geq 1.55$ & 0.359 & 0.625 & \textbf{0.719} & 2.0$\times$ & $3.0\times10^{-4}$ & 0.154 \\
64 & $R\geq 1.60$ & 0.109 & 0.406 & \textbf{0.547} & 5.0$\times$ & $4.2\times10^{-6}$ & 0.061 \\
64 & $R\geq 1.62$ & 0.078 & 0.281 & \textbf{0.328} & 4.2$\times$ & 0.001 & 0.304 \\
\bottomrule
\end{tabular}

\caption{Prespecified threshold-crossing rates on the 12,016-candidate grid. $B$ denotes query budget, Gain$_R$ is the ratio of \method{} and random-search hit rates, and $p_R/p_G$ are one-sided paired values against random search and GP-UCB. Reporting several thresholds prevents the discovery conclusion from depending on the single headline cutoff.}
\label{tab:supp-largegrid-thresholds}
\end{table*}

\begin{table*}[t]
\centering
\small
\setlength{\tabcolsep}{5pt}
\begin{tabular}{lrrrrrrr}
\toprule
$B$ & Random & GP-UCB & BasinLens & Gain$_R$ & $p_R$ & Gain$_G$ & $p_G$ \\
\midrule
16 & 16.89 & \textbf{16.58} & 16.67 & 0.22 & 0.188 & -0.09 & 0.746 \\
32 & 32.33 & \textbf{30.00} & 30.52 & 1.81 & 0.011 & -0.52 & 0.798 \\
64 & 61.27 & 51.69 & \textbf{49.69} & 11.58 & $2.4\times10^{-6}$ & 2.00 & 0.087 \\
\bottomrule
\end{tabular}

\caption{Mean capped first-hit step for search score at least $1.60$ on the largest PushT grid; lower is better. A run that does not cross the threshold is assigned one step beyond its budget. Gain columns are comparator minus \method{}, so positive values favor \method{}.}
\label{tab:supp-time-to-hit}
\end{table*}

\begin{figure}[t]
\centering
\includegraphics[width=.96\columnwidth]{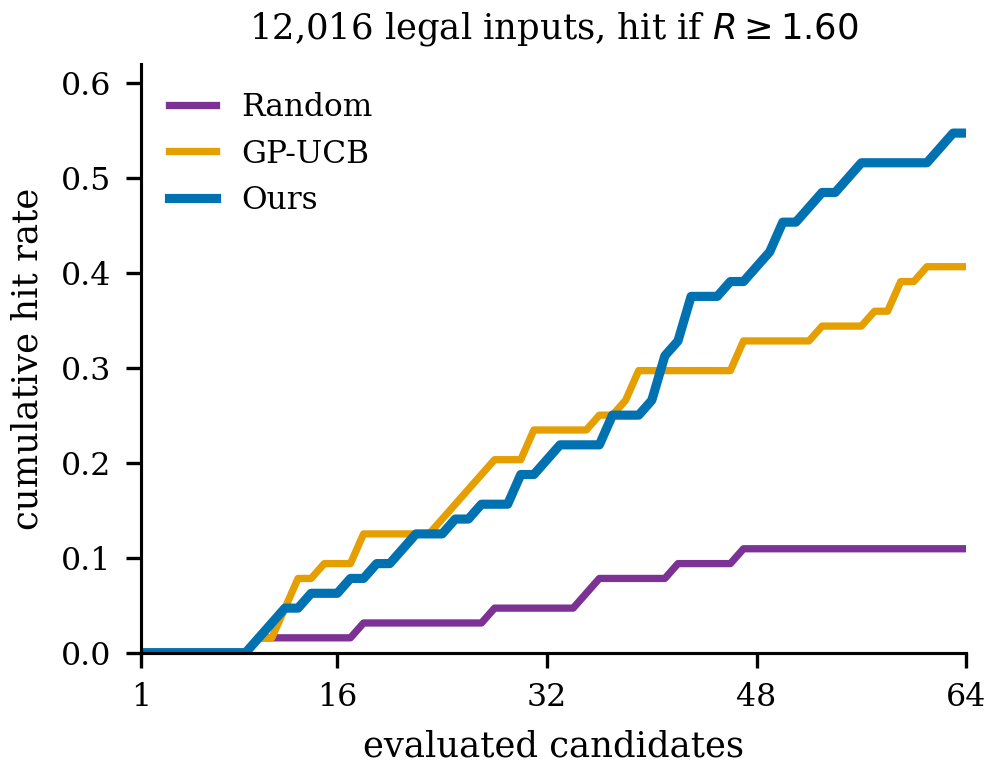}
\caption{Cumulative fraction of paired restarts that have discovered search score at least $1.60$ after each query on the 12,016-candidate PushT grid. The endpoint values are $0.547$ for \method{}, $0.406$ for GP-UCB, and $0.109$ for random search.}
\label{fig:supp-largegrid-cumulative}
\end{figure}

\begin{table}[!htbp]
\centering
\small
\setlength{\tabcolsep}{1.5pt}
\begin{tabular}{lrrrcc}
\toprule
Split & Random & GP-UCB & \method{} & $p_R$ & $p_G$ \\
\midrule
A & 1.388 & 1.393 & \textbf{1.433} & $7.81{\times}10^{-4}$ & 0.052 \\
B & 1.587 & 1.620 & \textbf{1.647} & $4.81{\times}10^{-9}$ & 0.010 \\
\bottomrule
\end{tabular}

\caption{Budget-32 best-search-score replication on two independent, non-overlapping seed splits of the same 3,510-candidate library (higher is better). Splits A and B use 64 and 128 paired restarts. $p_R$ and $p_G$ compare \method{} with random search and GP-UCB.}
\label{tab:supp-wide-search}
\end{table}

\begin{table}[!htbp]
\centering
\small
\setlength{\tabcolsep}{1.5pt}
\begin{tabular}{lrrrcc}
\toprule
Split & Random & Mid-score & Top & $p_R$ & $p_M$ \\
\midrule
A & 0.648 & 0.653 & \textbf{0.997} & 0.00458 & 0.00117 \\
B & 0.546 & 0.608 & \textbf{1.252} & $6.99{\times}10^{-4}$ & $7.77{\times}10^{-5}$ \\
\bottomrule
\end{tabular}

\caption{Fresh radius-1 neighborhood conditional value at risk (CVaR) around \method{} top-score anchors on the same two wide-grid splits (higher indicates greater prediction risk). $p_R$ and $p_M$ compare these neighborhoods with random-control and mid-score-control neighborhoods; this table does not compare \method{} neighborhoods with GP-UCB neighborhoods.}
\label{tab:supp-wide-neighborhood}
\end{table}

\section{Consolidated Quantitative Ledger}
The detailed tables above separate budgets, thresholds, and validation protocols. Table~\ref{tab:supp-consolidated-results} provides a single endpoint ledger for the principal online PushT searches so that their sample sizes and paired comparisons can be read without combining incompatible measurements. The mean-best rows measure the largest search score observed within a budget; the final two rows measure whether a run crossed the prespecified $1.60$ threshold. The latter comparison with GP-UCB is reported as descriptive evidence because its paired value is $p=0.061$.

\begin{table*}[t]
\centering
\small
\setlength{\tabcolsep}{3.5pt}
\begin{tabular}{lllrrrll}
\toprule
Protocol & Endpoint & Comparator & Comp. & \method{} & $\Delta$ & W/L/T & $p$ \\
\midrule
Online 6D, $B=32$ & Mean best $R$ & Random & 1.368 & \textbf{1.415} & \textbf{+0.047} & 67/27/2 & $2.25{\times}10^{-5}$ \\
Online 6D, $B=32$ & Mean best $R$ & GP-UCB & 1.375 & \textbf{1.415} & \textbf{+0.040} & 47/23/26 & 0.0028 \\
Wide 3,510, $B=32$ & Mean best $R$ & Random & 1.587 & \textbf{1.647} & \textbf{+0.060} & 95/31/2 & $4.81{\times}10^{-9}$ \\
Wide 3,510, $B=32$ & Mean best $R$ & GP-UCB & 1.620 & \textbf{1.647} & \textbf{+0.027} & 66/41/21 & 0.010 \\
Large 12,016, $B=64$ & Mean best $R$ & Random & 1.512 & \textbf{1.572} & \textbf{+0.060} & 49/14/1 & $5.56{\times}10^{-6}$ \\
Large 12,016, $B=64$ & Mean best $R$ & GP-UCB & 1.547 & \textbf{1.572} & \textbf{+0.025} & 37/16/11 & 0.0027 \\
Large 12,016, $B=64$ & Hit at $R{\geq}1.60$ & Random & 0.109 & \textbf{0.547} & \textbf{+0.438} & 34/6/24 & $4.18{\times}10^{-6}$ \\
Large 12,016, $B=64$ & Hit at $R{\geq}1.60$ & GP-UCB & 0.406 & \textbf{0.547} & \textbf{+0.141} & 18/9/37 & 0.061 \\
\bottomrule
\end{tabular}

\caption{Consolidated online LeWM PushT endpoints from existing run-level records. Comp. is the named comparator, $\Delta$ is \method{} minus that comparator, and W/L/T gives paired wins, losses, and ties. Rows remain protocol-specific and are not averaged across candidate libraries.}
\label{tab:supp-consolidated-results}
\end{table*}

The ledger also shows why discovery is not summarized by one universal performance number. Mean-best search score is useful for ranking methods within one library, while threshold crossing makes the rare-event discovery interpretation explicit. Held-out point risk and fresh-neighborhood risk are intentionally absent from this table because they are follow-up measurements on returned cases rather than additional online-search endpoints; those results remain in Tables~\ref{tab:supp-validation} and~\ref{tab:supp-wide-neighborhood}.

\section{Qualitative Failure Cases}
\begin{figure*}[t]
\centering
\includegraphics[width=.96\textwidth]{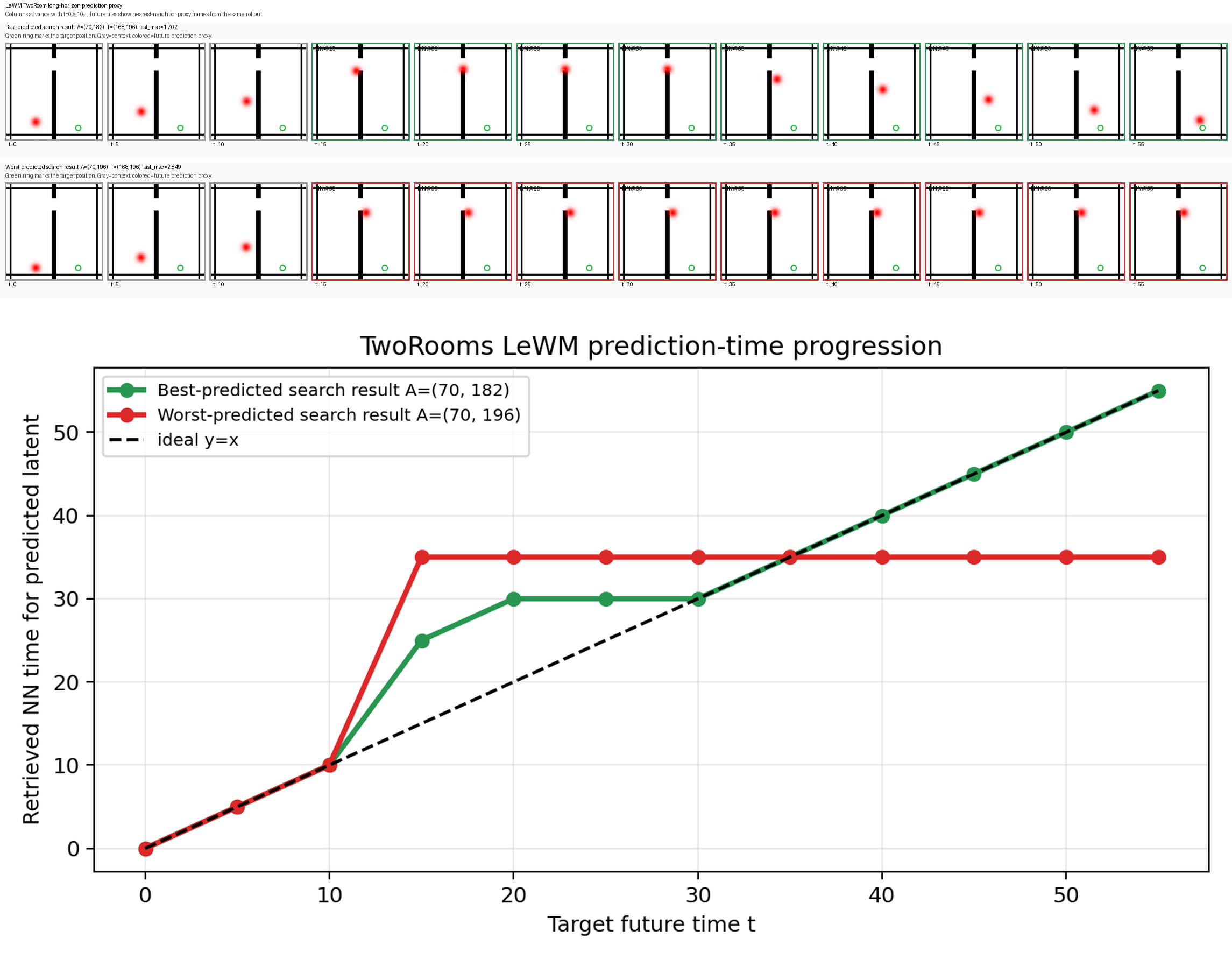}
\caption{TwoRooms case with a fixed target and two valid agent starts. The top panel compares matched environment frames with nearest-neighbor visual proxies for predicted future states. The lower curve maps requested future time to the retrieved prediction time. The selected high-error start (red) collapses to nearly the same predicted state after the early horizon, whereas the reference start (green) continues to track the requested future.}
\label{fig:supp-tworoom}
\end{figure*}

The deterministic TwoRooms example holds the target fixed and changes only the valid agent start. The two starts are close in the input grid, yet their predicted futures differ sharply. The reference start has future-position mean absolute error (MAE) $13.2$ pixels and final target-distance gap $27.2$ pixels; the selected start has MAE $89.0$ pixels and final gap $149.3$ pixels. Its latent-to-state probe MAE remains $1.66$ pixels, compared with $1.13$ pixels for the reference start. Figure~\ref{fig:supp-tworoom} visualizes the temporal mismatch: the selected prediction stops advancing after the early forecast steps even though the matched environment trajectory continues.

\begin{figure*}[t]
\centering
\includegraphics[width=.99\textwidth]{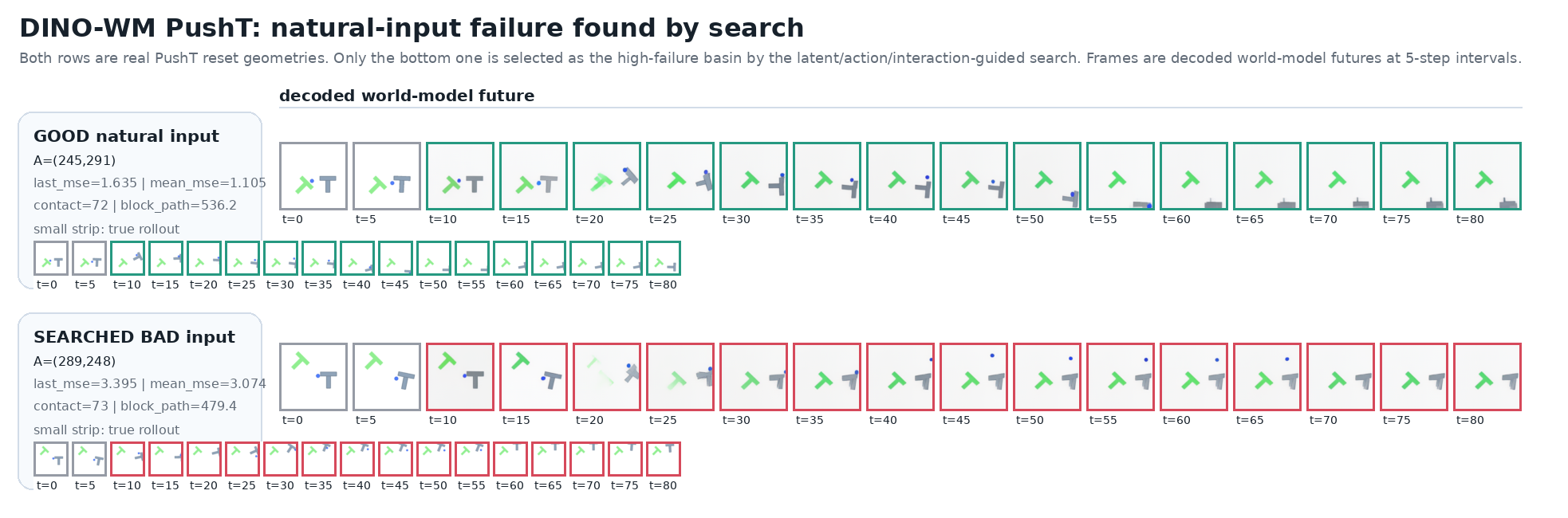}
\caption{Illustrative DINO-WM PushT discovery record. Each large strip shows decoded world-model futures at five-step intervals, and each small strip shows the matched environment rollout. The lower legal input, $A=(289,248)$, was returned with higher visual-feature prediction risk than the displayed comparison input, $A=(245,291)$. This case illustrates the content of a returned record and is not an additional independent search trial.}
\label{fig:supp-dino-pusht-case}
\end{figure*}

Figure~\ref{fig:supp-dino-pusht-case} provides a second-model PushT example using DINO-WM rather than the LeWM state probe. Both rows are valid reset geometries with active matched rollouts. The displayed searched case has final-step visual-feature MSE $3.395$ and mean horizon MSE $3.074$, compared with $1.635$ and $1.105$ for the displayed lower-risk case. Their matched rollouts contain 73 and 72 contact steps and block paths of $479.4$ and $536.2$ pixels, respectively, so the higher discrepancy is not produced by an inactive rollout.

\begin{figure}[t]
\centering
\includegraphics[width=.98\columnwidth]{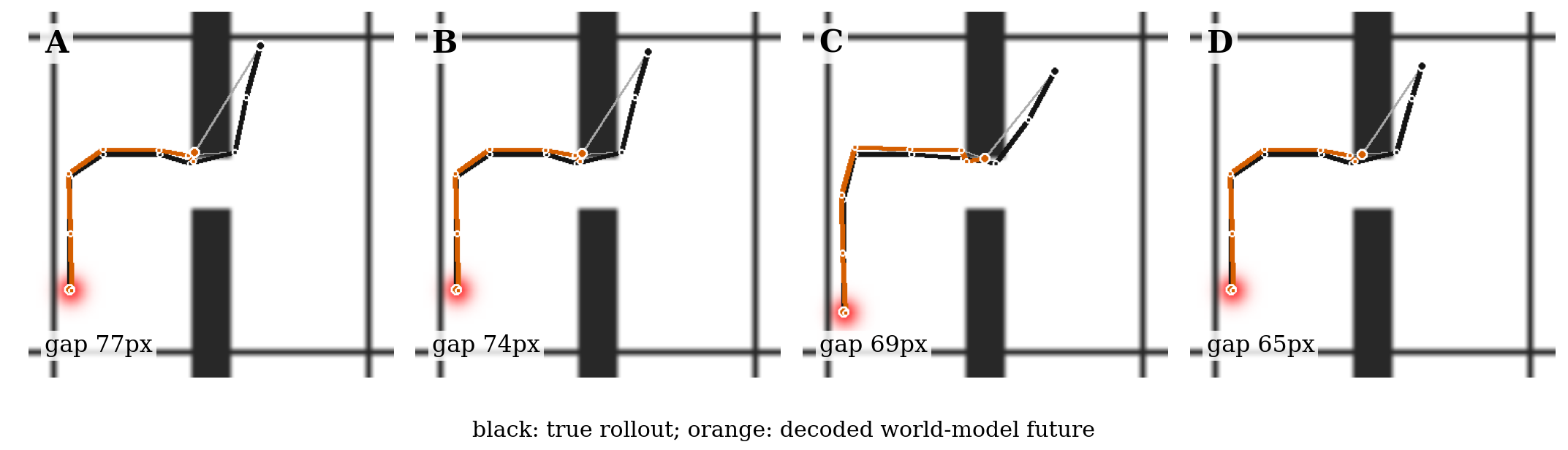}
\caption{Four valid DINO-WM Wall configurations from the shifted local library. Black shows the matched environment rollout and orange the decoded world-model future. In each case, the environment trajectory passes through the opening while the predicted future remains on the original side. The gap labels denote the opening widths in pixels.}
\label{fig:supp-wall}
\end{figure}

A historical DINO-WM Wall scan supplies a visually direct example of the same problem at a different prediction interface. The shifted library contains 900 prioritized local start, goal, and door configurations drawn from 7,560 Cartesian combinations. Ninety-two inputs satisfy the prespecified visible no-cross criterion: the environment trajectory crosses the opening, while the decoded predicted future does not. Figure~\ref{fig:supp-wall} shows four cases. This result is used as a scoped qualitative failure bank, not as a distribution-wide frequency estimate or a claim that the canonical \method{} variant dominates search baselines on Wall.

\begin{figure*}[t]
\centering
\includegraphics[width=.98\textwidth]{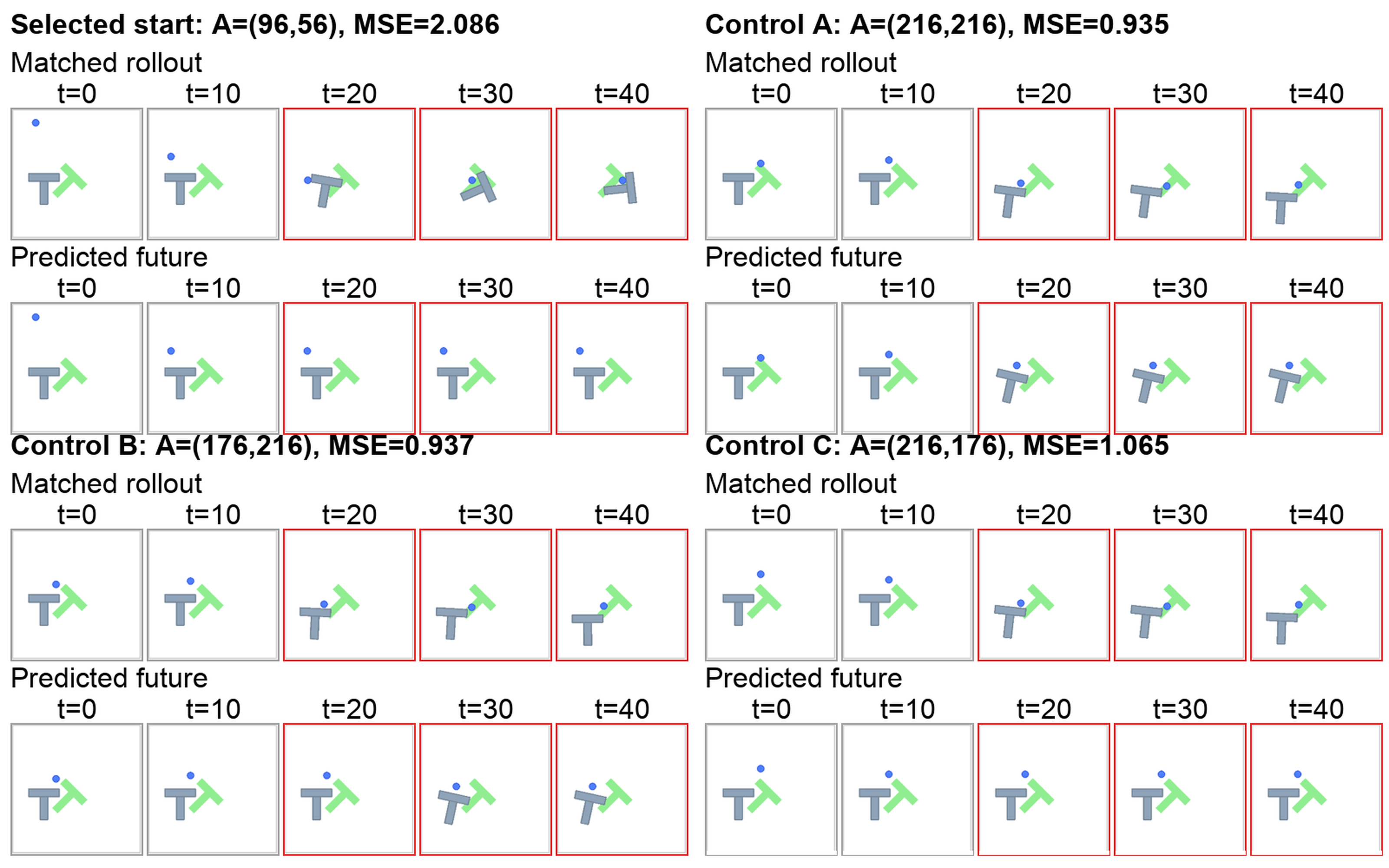}
\caption{PushT temporal comparison under fixed block--goal geometry. Each pair of rows shows the matched environment rollout and nearest-neighbor visual proxies for the predicted future. The selected start $A=(96,56)$ has the largest displayed final-step MSE and repeatedly maps later target times to the same early predicted state. The comparison starts show lower error and more temporal variation.}
\label{fig:supp-pusht-multicase}
\end{figure*}

Figure~\ref{fig:supp-pusht-multicase} shows one selected PushT start and three comparison starts under the same fixed block--goal geometry. The selected start has final-step MSE $2.086$, compared with $0.935$, $0.937$, and $1.065$ for the displayed comparisons. Its nearest-neighbor prediction proxy repeatedly retrieves the same early state, while the matched environment rollout continues to move the block. The panel is qualitative; the held-out and neighborhood protocols provide the independent quantitative evidence.

\subsection{What the Discovered Cases Reveal}
The qualitative cases expose two recurring forms of error that an average benchmark score does not localize. In the temporal cases, requested later futures map to an early or weakly changing predicted state even though the environment continues to evolve. In the topological case, the predicted future remains on the wrong side of a traversable opening. Table~\ref{tab:supp-failure-characterization} records the changed legal factor, the visible failure pattern, and the evidence boundary for each case family.

\begin{table*}[t]
\centering
\small
\setlength{\tabcolsep}{4pt}
\begin{tabularx}{\textwidth}{@{}p{1.05in}p{1.20in}X X@{}}
\toprule
Case family & Legal factor varied & Observed prediction pattern & Evidence boundary \\
\midrule
LeWM TwoRooms & Agent start with fixed target & Predicted position stops advancing and later requested times retrieve nearly the same early state & Deterministic paired visualization; not a frequency estimate over all navigation layouts \\
LeWM PushT & Agent start with fixed block and goal & Later target times repeatedly map to an early state while the matched rollout continues moving the block & Illustrated case; independent held-out and neighborhood tables support repeatability at the protocol level \\
DINO-WM PushT & Legal agent, block, and goal geometry & Decoded future diverges from the active matched rollout and yields elevated visual-feature error & Illustrative returned record; quantitative search evidence comes from the paired replay tables \\
DINO-WM Wall & Start, goal, and opening geometry & Environment trajectory crosses the opening while the decoded prediction remains on the original side & 92 visible no-cross cases in the prioritized 900-input shifted library; not a distribution-wide prevalence claim \\
\bottomrule
\end{tabularx}
\caption{Qualitative characterization of discovered legal-input failures. The cases explain what fails while keeping each conclusion attached to the protocol that produced it.}
\label{tab:supp-failure-characterization}
\end{table*}

\section{Validation and Planning}
The seed-matched model predictive control (MPC) protocol uses the four validated inputs with the highest held-out prediction CVaR. Each case is evaluated with context seeds 0--7. After a 15-step observed prefix, the controllers replan every 10 simulator steps for 40 control steps using a 30-step horizon. At each replan, both controllers use proposal seeds 100--115 with the same state-conditioned PushT proposal generator (\texttt{WeakPolicy}) to instantiate 16 proposals from their respective current states. The model-ranked MPC controller scores model-predicted state sequences, whereas the environment-oracle-ranked MPC controller scores simulator state sequences, using
\[
U=-g+0.8p+0.3m+0.2u+0.2c .
\]
Here, $g$ is the final block--goal distance divided by its distance at the forecast origin; $p=1-g$ is normalized block--goal progress; $m$ is the fraction of forecast steps whose block--goal distance decreases by at least 2 pixels from the immediately preceding state; $u$ is the fraction of forecast steps whose agent--block distance is at most 90 pixels and whose block moves at least 8 pixels from the immediately preceding state; and $c$ is the fraction of forecast steps satisfying the stepwise conditions for both $m$ and $u$. Each controller selects the proposal with maximum $U$ under its ranking rule, executes its first 10 actions, and then replans from its resulting state. A random controller samples a proposal generated from the same proposal-seed set. If a simulated proposal terminates before the fixed horizon, its last state is repeated and zero actions provide absorbing padding.

The stratified analysis uses 24 proposal seeds, 100--123, with the same context seeds and replanning schedule. Its top group contains the four task-active inputs with the highest held-out prediction risk among the \method{} top-eight anchors and their radius-1 neighbors; the mid-score and random-control groups contain the first four task-active inputs in their prespecified anchor lists. This fixed, task-active set supports within-set controller comparisons rather than a prevalence estimate.

\begin{table*}[!htbp]
\centering
\small
\setlength{\tabcolsep}{1.5pt}
\begin{tabular}{lllrrrl}
\toprule
Protocol & Endpoint & Control & Control value & Selected & $\Delta$ & Evidence \\
\midrule
Fixed PushT & Final BG ratio & Reference start & 0.941 & \textbf{1.110} & \textbf{+0.169} & Higher in 98.4\% of 64 pairs \\
Held-out point & CVaR & Default reset & 1.186 & \textbf{1.311} & \textbf{+0.125} & 24 reevaluated cases \\
Wide-grid local & Radius-1 CVaR & Random controls & 0.546 & \textbf{1.252} & \textbf{+0.706} & $p=6.99{\times}10^{-4}$ \\
Wide-grid local & Radius-1 CVaR & Mid-score controls & 0.608 & \textbf{1.252} & \textbf{+0.644} & $p=7.77{\times}10^{-5}$ \\
\bottomrule
\end{tabular}

\caption{Fixed-case, held-out point, and fresh-neighborhood evidence. BG denotes block--goal and CVaR denotes conditional value at risk. For the BG ratio, higher means less predicted task progress; for CVaR, higher means greater prediction risk. Here, $\Delta$ is Selected minus Control.}
\label{tab:supp-validation}
\end{table*}

\begin{table*}[!htbp]
\centering
\small
\setlength{\tabcolsep}{1.5pt}
\begin{tabular}{llrrrll}
\toprule
Protocol & Context group & Model-ranked & Env.-oracle-ranked & $\Delta$ & W/L/T & $p$ \\
\midrule
MPC, 16 proposal seeds & Selected & -0.080 & \textbf{0.445} & \textbf{+0.525} & 30/2/0 & $1.23{\times}10^{-7}$ \\
Stratified MPC & Top held-out risk & -0.100 & \textbf{0.344} & \textbf{+0.443} & 29/3/0 & $1.28{\times}10^{-6}$ \\
Stratified MPC & Mid-control & 0.061 & \textbf{0.532} & \textbf{+0.471} & 29/3/0 & $1.28{\times}10^{-6}$ \\
Stratified MPC & Random-control & 0.045 & \textbf{0.373} & \textbf{+0.328} & 28/4/0 & $9.65{\times}10^{-6}$ \\
\bottomrule
\end{tabular}

\caption{Seed-matched model predictive control (MPC) results. At every replan, the model-ranked and environment-oracle-ranked controllers use the same proposal-seed set and state-conditioned proposal generator; higher environment-side progress is better. Here, $\Delta$ is environment-oracle-ranked minus model-ranked progress. W/L/T denotes paired wins/losses/ties for Oracle versus Model, and $p$ is the one-sided exact sign-test value.}
\label{tab:supp-consequence}
\end{table*}

\begin{table}[!htbp]
\centering
\small
\setlength{\tabcolsep}{1pt}
\begin{tabularx}{\columnwidth}{@{}rrrrr>{\raggedright\arraybackslash}X@{}}
\toprule
Seeds & Model & Oracle & Random & Gap & Paired evidence \\
\midrule
8 & -0.107 & 0.253 & -0.072 & 0.360 & 24/32 wins; $p=0.0035$; 11 sign reversals \\
16 & -0.080 & 0.445 & -0.274 & 0.525 & 30/32 wins; $p=1.23{\times}10^{-7}$; 16 sign reversals \\
\bottomrule
\end{tabularx}

\caption{MPC sensitivity to proposal-seed set size. Model, Oracle, and Random are environment-side progress under the model-ranked, environment-oracle-ranked, and random controllers. Gap is Oracle minus Model; higher progress is better. Paired evidence reports Oracle wins, the one-sided exact sign-test value, and the number of negative-to-positive sign reversals.}
\label{tab:supp-mpc-library}
\end{table}

\section{Statistical Comparisons}
Search methods share the same algorithm-restart seed within each comparison, so the unit is the paired run. For a directional comparison, let $n_+$ and $n_-$ be the numbers of strictly positive and negative paired differences; exact sign-test values are upper-tail binomial probabilities under success probability $1/2$, with ties removed. Mean differences and standard errors are still computed from all paired run-level values. This test does not assume normally distributed search scores.

Threshold crossing is binary. Let $b$ count restarts in which \method{} crosses the threshold but the comparator does not, and let $c$ count the reverse. The one-sided exact McNemar value is the binomial upper tail for $b$ successes among $b+c$ discordant pairs. Threshold-averaged discovery first averages the 16 prespecified hit indicators within each run and then applies the paired scalar comparison to those run-level averages.

The fresh-neighborhood comparison treats the eight anchor-level neighborhood means in each group as its units. For two groups of eight, the exact permutation distribution enumerates all $\binom{16}{8}$ assignments of the 16 observed anchor statistics into two groups of size eight. The reported one-sided value is the fraction whose difference in group means is at least the observed difference. Candidate overlap across different anchor neighborhoods is retained because the inferential unit is the prespecified anchor, not the unique candidate.

\section{Component Isolation and DINO-WM Replication}
The component analysis separates global uncertainty, coordinate expansion, pairwise features, and multi-anchor expansion. The DINO-WM replay then compares continuation policies after an identical random scout, so differences in best search score and reference-component coverage arise only from the remaining queries. All DINO-WM replay tables use 128 paired replay seeds, 0--127.
\begin{table}[!htbp]
\centering
\small
\setlength{\tabcolsep}{1pt}
\begin{tabular}{lrrr}
\toprule
Method & Hit@8 $\uparrow$ & Hit@16 $\uparrow$ & Hit@32 $\uparrow$ \\
\midrule
Random & 0.361 & 0.562 & 0.828 \\
GP-UCB & 0.475 & 0.689 & 0.801 \\
CEM & 0.326 & 0.572 & 0.811 \\
\frontier{} & 0.287 & 0.527 & 0.773 \\
\method{} & 0.455 & 0.672 & 0.803 \\
\interaction{} & \textbf{0.512} & \textbf{0.869} & 0.961 \\
\beam{} & 0.479 & 0.834 & \textbf{0.963} \\
\bottomrule
\end{tabular}

\caption{Top-5\% hit rate on the 431-candidate cached combinatorial grid across query budgets and 512 paired restarts. CEM denotes the cross-entropy method. For \interaction{}/\beam{} versus GP-UCB, one-sided exact sign-test values are $0.083/0.472$ at budget 8, $1.81{\times}10^{-16}/2.10{\times}10^{-9}$ at budget 16, and $7.32{\times}10^{-17}/8.77{\times}10^{-20}$ at budget 32.}
\label{tab:supp-method-budget}
\end{table}

\begin{table*}[!htbp]
\centering
\small
\setlength{\tabcolsep}{2pt}
\begin{tabular}{@{}rrlrrrr@{}}
\toprule
$N$ & Top & Method & Best score $\uparrow$ & Top count $\uparrow$ & Comp. $\uparrow$ & Hit $\uparrow$ \\
\midrule
512 & 5\% & Random & 2.712 & 1.594 & 0.984 & 0.812 \\
 &  & GP-UCB & 2.863 & 5.617 & 3.195 & 1.000 \\
 &  & \frontier{} & 2.849 & 5.781 & \textbf{4.914} & 0.906 \\
 &  & \method{} & \textbf{2.908} & \textbf{7.023} & 4.898 & 1.000 \\
\midrule
384 & 5\% & Random & 2.748 & 1.688 & 1.148 & 0.828 \\
 &  & GP-UCB & 2.845 & 4.062 & 3.180 & 0.977 \\
 &  & \frontier{} & 2.840 & \textbf{4.773} & \textbf{4.344} & 0.992 \\
 &  & \method{} & \textbf{2.858} & 4.734 & 4.195 & 0.992 \\
\bottomrule
\end{tabular}


\caption{DINO World Model (DINO-WM) PushT common-scout replication on independent 512- and 384-candidate libraries. $N$ is the candidate-library size. Every method receives the same eight random scout points and a total budget of 32. Best score measures point discovery; Top count is the mean number of evaluated nodes in the prespecified top-score set; Comp.\ is the mean reference-component coverage $C_q$; and Hit is the fraction of runs that evaluate at least one top-score candidate. Higher is better.}
\label{tab:supp-dino-replication}
\end{table*}

\begin{table*}[!htbp]
\centering
\small
\setlength{\tabcolsep}{5pt}
\begin{tabular}{llrrrr}
\toprule
Library & Outcome & $\Delta_G$ & $p_G$ & $\Delta_F$ & $p_F$ \\
\midrule
512 candidates & Best score & 0.045 & $3.1{\times}10^{-4}$ & -- & -- \\
 & Comp. cov. & 1.703 & $1.1{\times}10^{-14}$ & -0.016 & 0.929 \\
\midrule
384 candidates & Best score & 0.013 & 0.006 & -- & -- \\
 & Comp. cov. & 1.016 & $6.9{\times}10^{-6}$ & -0.148 & 0.988 \\
\bottomrule
\end{tabular}

\caption{Paired DINO-WM top-5\% comparisons for \method{}. Comp.\ denotes reference-component coverage, $\Delta_G=\method{}-\mathrm{GP\text{-}UCB}$, and $\Delta_F=\method{}-\frontier{}$. Each $p$ entry is the corresponding one-sided exact sign-test value. The frontier-only variant has slightly higher mean component coverage in both libraries.}
\label{tab:supp-dino-paired}
\end{table*}

\begin{table}[!htbp]
\centering
\small
\setlength{\tabcolsep}{2pt}
\begin{tabular}{@{}llrrrr@{}}
\toprule
$(B,S)$ & Top & Best score & $\Delta_G$ & & \\
\midrule
$(16,8)$ & 10\% & 2.738 & 0.008 \\
$(16,8)$ & 5\% & 2.738 & 0.008 \\
$(32,4)$ & 10\% & 2.880 & 0.024 \\
$(32,4)$ & 5\% & 2.880 & 0.024 \\
$(32,8)$ & 10\% & 2.908 & 0.045 \\
$(32,8)$ & 5\% & 2.908 & 0.045 \\
$(32,16)$ & 10\% & 2.877 & 0.023 \\
$(32,16)$ & 5\% & 2.877 & 0.023 & & \\
\midrule
$(B,S)$ & Top & Comp. & $\Delta_G$ & $p_G$ & $\Delta_F$ \\
\midrule
$(16,8)$ & 10\% & 2.891 & 0.750 & $1.9{\times}10^{-7}$ & -0.945 \\
$(16,8)$ & 5\% & 1.992 & 0.547 & $1.9{\times}10^{-5}$ & -0.828 \\
$(32,4)$ & 10\% & 7.141 & 1.961 & $1.9{\times}10^{-7}$ & -0.117 \\
$(32,4)$ & 5\% & 5.016 & 1.578 & $1.3{\times}10^{-10}$ & -0.164 \\
$(32,8)$ & 10\% & 6.891 & 2.320 & $9.5{\times}10^{-15}$ & 0.094 \\
$(32,8)$ & 5\% & 4.898 & 1.703 & $1.1{\times}10^{-14}$ & -0.016 \\
$(32,16)$ & 10\% & 5.602 & 1.430 & $2.4{\times}10^{-14}$ & -1.398 \\
$(32,16)$ & 5\% & 4.055 & 1.117 & $6.0{\times}10^{-14}$ & -1.148 \\
\bottomrule
\end{tabular}

\caption{DINO-WM sensitivity over total query budget $B$ and scout size $S$. Comp.\ denotes reference-component coverage, $\Delta_G=\method{}-\mathrm{GP\text{-}UCB}$, and $\Delta_F=\method{}-\frontier{}$. The upper panel reports point discovery; the lower panel reports component coverage, with $p_G$ giving the one-sided exact sign-test value for the comparison with GP-UCB.}
\label{tab:supp-dino-sensitivity}
\end{table}

\section{Cross-Model Scope}
This section reports the JEPA-WM, DIAMOND, and IRIS protocols. Model names are used directly rather than repeatedly expanding branded acronyms.
\begin{table*}[t]
\centering
\small
\setlength{\tabcolsep}{4pt}
\begin{tabularx}{\textwidth}{@{}>{\raggedright\arraybackslash}Xrrrr>{\raggedright\arraybackslash}p{1.28in}@{}}
\toprule
Model, case, and metric & $B/N$ & Random & UCB & Structured & Paired $p$ \\
\midrule
LeWM PushT 6D, best search score & 32/3510 & 1.587 & 1.620 & \textbf{1.647} & $4.81{\times}10^{-9}$/0.010 \\
DIAMOND Breakout, best MSE ($\times10^{-4}$) & 16/128 & 1.6930 & \textbf{2.3203} & 1.8154 & 1.000/1.000 \\
DIAMOND Pong, best MSE ($\times10^{-4}$) & 16/128 & \textbf{2.3478} & 2.0902 & 2.2117 & 0.500/1.000 \\
IRIS Breakout, best MSE ($\times10^{-5}$) & 6/36 & 1.4400 & \textbf{1.4401} & 1.4400 & --/1.000 \\
IRIS Pong, best MSE ($\times10^{-4}$) & 6/36 & 1.1885 & \textbf{1.2057} & 1.0004 & 1.000/1.000 \\
JEPA-WM PointMaze, composite latent risk & 16/480 & 4.935 & 4.042 & \textbf{5.113} & \shortstack[l]{$8.55{\times}10^{-4}$/\\$2.49{\times}10^{-8}$} \\
\bottomrule
\end{tabularx}
\caption{Cross-model search summary. $B/N$ is query budget/library size. Structured denotes \method{} for LeWM and the protocol-specific policies described in this section for the other models. The final column gives paired $p$ values versus random/UCB; a dash marks an unavailable comparison. Higher is better, and bold marks the largest score in each row.}
\label{tab:cross-worldmodel-search-compact}
\end{table*}

The JEPA-WM PointMaze library contains $480=12{\times}10{\times}4$ inputs: reset seeds 0--11, ten valid two-dimensional action templates, four scales $\{0.25,0.5,0.75,1.0\}$, and a fixed three-action horizon. The objective is visual-latent mean squared error plus $0.1$ times proprioceptive-latent mean squared error. The templates are right $(1,0)$, left $(-1,0)$, up $(0,1)$, down $(0,-1)$, the diagonals $(0.75,0.75)$ and $(0.75,-0.75)$, and the two-step sequences $[(1,0),(-1,0)]$, $[(0,1),(0,-1)]$, $[(1,0),(0,1)]$, and $[(0,1),(-1,0)]$. Multi-action templates cycle to length three, are scaled and clipped to $[-1,1]^2$, and repeat each macro action for five simulator steps. The library order is shuffled once with seed 13, and results average 128 paired search seeds, 0--127. The search representation is
\[
\begin{aligned}
\phi(x)=\bigl[&s_{\mathrm{seed}}/64,\bar a_x,\bar a_y,\|\bar a\|_2,\sigma_x+\sigma_y,\\
&\operatorname{mean}_t\|a_{t+1}-a_t\|_2,\gamma\bigr].
\end{aligned}
\]
Here, $s_{\mathrm{seed}}$ is the reset seed, $\gamma$ is the action scale, $\bar a$ is the mean action vector, $\sigma_x$ and $\sigma_y$ are its componentwise temporal standard deviations, and the penultimate entry is the mean change between successive actions.
For observations $(x_i,r_i)$, the kernel-weighted UCB comparator uses weights $w_i(x)=\exp(-\|\phi(x)-\phi(x_i)\|_2^2/(2\cdot0.55^2))$, their weighted risk mean $\mu$ and variance $v$, and acquisition $A(x)=\mu(x)+\sqrt{v(x)+\eta^2/(\sum_iw_i+10^{-6})}$, where $\eta=\max\{\operatorname{std}(r),0.25\max_i|r_i|,10^{-8}\}$. Within each paired search seed, all methods share the first two shuffled scout candidates. The global comparator maximizes $A$ over every unseen input. The structured policy instead maximizes $A$ over neighbors of the three highest-prediction-risk observations, reverting to all unseen inputs only if that frontier is empty. A neighbor shares the reset seed and satisfies at least one of $|\gamma'-\gamma|\leq0.26$, $\|\bar a'-\bar a\|_2\leq0.85$, or $\bar a'^\top\bar a>0$.

\begin{table*}[t]
\centering
\small
\setlength{\tabcolsep}{5pt}
\begin{tabular}{rrlrrrr}
\toprule
$B$ & $N$ & Method & Mean best risk & Top-5 hit & Top-10 hit & Top-10\% hit \\
\midrule
4 & 480 & Random & 3.635 & 0.016 & 0.031 & 0.281 \\
  &     & GP-UCB & 3.449 & 0.008 & 0.016 & 0.203 \\
  &     & Structured & \textbf{3.902} & \textbf{0.039} & \textbf{0.094} & \textbf{0.344} \\
\midrule
8 & 480 & Random & \textbf{4.281} & 0.062 & 0.133 & 0.555 \\
  &     & GP-UCB & 3.655 & 0.008 & 0.047 & 0.414 \\
  &     & Structured & 4.226 & \textbf{0.086} & \textbf{0.156} & \textbf{0.648} \\
\midrule
16 & 480 & Random & 4.935 & 0.125 & 0.266 & 0.781 \\
   &     & GP-UCB & 4.042 & 0.008 & 0.141 & 0.602 \\
   &     & Structured & \textbf{5.113} & \textbf{0.234} & \textbf{0.359} & \textbf{0.867} \\
\midrule
32 & 480 & Random & 5.907 & 0.297 & 0.484 & 0.961 \\
   &     & GP-UCB & 5.186 & 0.008 & 0.383 & \textbf{0.992} \\
   &     & Structured & \textbf{6.277} & \textbf{0.359} & \textbf{0.570} & 0.922 \\
\bottomrule
\end{tabular}

\caption{JEPA-WM PointMaze budget sweep over 128 paired replay seeds. Higher is better for all columns because the objective is to discover inputs with higher prediction risk. The structured policy is strongest in mean best risk at budgets 4, 16, and 32; random search is slightly higher at budget 8, illustrating that method ordering depends on the available budget and risk landscape.}
\label{tab:supp-jepa-budget}
\end{table*}

DIAMOND inputs are a reset seed and four valid action-prefix entries at fixed context index four; IRIS inputs are a reset seed, pre-context no-op count, and action-template label at a fixed context length and a prediction horizon of four. The final DIAMOND studies use candidate-library size $N=128$, query budget $B=16$, and eight paired search seeds; the IRIS studies use $N=36$, $B=6$, and four paired search seeds. Both optimize future-frame MSE. Their portfolio method follows the random baseline's candidate stream until at least three risks have been observed and
\[
\begin{aligned}
r_{\max}>{}&\operatorname{median}(r)\\
&+\max\{6\operatorname{MAD}(r),\\
&\qquad 0.5|\operatorname{median}(r)|,10^{-8}\},
\end{aligned}
\]
where $\operatorname{MAD}$ is the median absolute deviation of the observed risks.
The condition is recomputed after every query. When it is active, the first query is structured and the portfolio then alternates one structured query with one query from the random stream; when it is inactive, it continues the random stream. An exhausted random stream falls back to structured selection. DIAMOND forms typed replacements of reset seed or one action-prefix entry around the three highest-prediction-risk observations and adds the eight leading global UCB proposals; IRIS replaces seed, no-op count, or template around the two highest-prediction-risk observations and adds the five leading global UCB proposals. Their kernel-weighted UCB rules use exploration coefficient 1.0 and length scales 0.75 and 0.60, respectively, and query the candidate with the largest score.

\section{Reproduction Record}
Each reported protocol can be reconstructed from five record types. Keeping these records separate makes it possible to verify candidate validity, replay search decisions, and recompute summaries without rerunning model inference.

\begin{table*}[t]
\centering
\small
\setlength{\tabcolsep}{5pt}
\begin{tabularx}{\textwidth}{@{}p{1.15in}X X@{}}
\toprule
Record & Required fields & Verification supported \\
\midrule
Candidate library & Stable identifier, typed input values, normalized representation, validity status & Candidate count, coordinate ranges, one-coordinate replacement graph \\
Evaluation record & Candidate identifier, environment seeds, per-seed losses, activity measurements, prediction risk, search score & Risk aggregation, activity adjustment, equality of cached values across methods \\
Search trace & Restart seed, ordered scout, ordered queries, acquisition values, top-$k$ output & Query budget, no repeated query, common-scout pairing, best-score and hit-rate summaries \\
Validation record & Selected identifier, disjoint validation seeds, per-seed losses, control identifier & Held-out point comparison without acquisition leakage \\
Neighborhood record & Anchor identifier, generated radius-1 identifiers, group label, fresh losses & Neighborhood validity, anchor-level aggregation, exact permutation test \\
\bottomrule
\end{tabularx}
\caption{Minimal record schema for reproducing the reported discovery and validation results. Tables and figures are derived from these records rather than transcribed from console output.}
\label{tab:supp-record-schema}
\end{table*}

\subsection{Executable Code Map}
The code supplement is organized by evidence type rather than by paper section. Table~\ref{tab:supp-code-map} maps the principal reported protocols to their retained entry points. Some filenames retain the historical string \texttt{basin\_gcg}; this is an implementation filename only and does not denote gradient-based prompt optimization.

\begin{table*}[t]
\centering
\small
\setlength{\tabcolsep}{4pt}
\begin{tabularx}{\textwidth}{@{}p{1.38in}X X@{}}
\toprule
Evidence type & Entry point & Primary record produced or consumed \\
\midrule
LeWM PushT discovery & \path{scripts/lewm_pusht_online_basin_gcg.py} & Valid candidate library, per-candidate evaluations, and ordered search traces \\
Held-out point validation & \path{scripts/validate_online_basin_candidates.py} & Selected identifiers and disjoint-seed point-validation records \\
Fresh-neighborhood persistence & \path{scripts/validate_online_widegrid_neighborhoods.py} & Radius-1 candidate lists, fresh losses, and anchor-level summaries \\
Component isolation & \path{scripts/combinatorial_basin_gcg_benchmark.py} & Cached paired search traces for the global, frontier, interaction, and beam policies \\
LeWM state readout & \path{scripts/lewm_pusht_state_probe_mainline.py} & Probe checkpoint, held-out errors, and decoded state sequences \\
Controller consequence & \path{scripts/lewm_pusht_closed_loop_mpc_audit.py} & Seed-matched proposal rankings and environment-side controller outcomes \\
DINO-WM PushT & \path{scripts/dino_pusht_online_basin_gcg.py} & Visual-feature risk evaluations and discovery traces \\
DINO-WM common-scout replay & \path{scripts/dino_basin_expansion_audit.py} & Paired continuation-policy traces and component-coverage summaries \\
JEPA-WM PointMaze & \path{scripts/jepa_wm_pointmaze_search.py} & Latent-risk candidate records and budget-sweep traces \\
Atari interfaces & \path{scripts/diamond_atari_multiseed_truepred.py}; \path{scripts/iris_atari_search_smoke.py} & Future-frame risk records for DIAMOND and IRIS \\
\bottomrule
\end{tabularx}
\caption{Code-to-evidence map for the anonymous supplement. Each entry point accepts explicit upstream-repository and checkpoint paths rather than embedding machine-local locations.}
\label{tab:supp-code-map}
\end{table*}

The release uses \texttt{uv}, not Conda. Running \texttt{uv sync --frozen} under Python 3.12 reconstructs the locked Python environment from \texttt{pyproject.toml} and \texttt{uv.lock}; \texttt{uv run python -m compileall -q scripts} provides a dependency-light syntax check of every retained entry point. Model-specific execution additionally requires the cited upstream repositories and checkpoints. Those assets remain under their original licenses, so the adapters receive their locations through explicit command-line arguments or configuration fields rather than anonymous-repository-relative assumptions.

A complete rerun first constructs and validates the finite candidate library, then generates evaluation records, replays or executes paired search traces, and finally runs held-out point or neighborhood evaluation from frozen selected identifiers. Tables are aggregated only after these records pass the consistency checks below. This order prevents a validation result from changing the candidate library, search trace, or shortlist that it is intended to assess.

Before aggregation, every trace is checked for a query count no larger than $B$, unique candidate identifiers within a restart, and membership of every query in the protocol's valid library. Common-scout comparisons additionally verify identical scout identifiers and values across methods. Held-out records are checked for disjoint search and validation seed sets. Neighborhood records are regenerated from typed coordinates and compared with their stored candidate identifiers; duplicate candidates within one anchor neighborhood are removed, while a candidate shared by two different anchors remains in both anchor-level statistics. Finally, all displayed means, differences, and test values are recomputed from the unrounded records.

The main search curves use cumulative threshold indicators from complete search traces. Validation bars use candidate-level or anchor-level measurements from separate records, as stated in each caption. The TwoRooms and Wall panels are generated from matched environment and model-prediction outputs for the displayed valid inputs. They are illustrative cases and are not used as additional independent samples in the statistical comparisons.

\section{Interpretation of Cross-Model Results}
The cross-model rows establish that natural-input failure discovery can be instantiated whenever a world model exposes an executable input specification and a prediction target that can be compared with the environment. They do not define a universal numerical risk scale: LeWM, DINO-WM, JEPA-WM, DIAMOND, and IRIS operate in different output spaces, so magnitudes are interpreted only within each protocol. Likewise, the structured policy is not assumed to dominate every comparator. The DINO-WM and JEPA-WM studies show positive cases for typed expansion, whereas the Atari rows document settings in which random search or the UCB comparator is stronger. These outcomes are retained because the central object is the discovered prediction failure and its input condition, not a cross-architecture optimizer leaderboard.

The main validation claim remains tied to LeWM PushT, where selected points and neighborhoods are evaluated with disjoint seeds. The DINO-WM replication uses two independent candidate libraries and common-scout replay, providing method-level replication but not the same held-out neighborhood protocol. The TwoRooms and Wall cases provide interpretable views of temporal and topological prediction failure. This separation keeps each result attached to the protocol that generated it.

\section{Reporting Natural-Input Failure Discovery}
A complete discovery result should identify the input condition, the prediction quantity that failed, and the evidence that the case is not an artifact of invalid inputs or reused evaluation noise. The search algorithm is only one part of that record. Table~\ref{tab:supp-reporting} summarizes the reporting fields used in this paper and their purpose.

\begin{table*}[t]
\centering
\small
\setlength{\tabcolsep}{5pt}
\begin{tabularx}{\textwidth}{@{}p{1.12in}X X@{}}
\toprule
Reporting item & Required content & Question answered \\
\midrule
Input specification & Typed coordinates, allowed values, and environment-level validity checks & Is the discovered condition executable without an adversarial or off-support perturbation? \\
Prediction interface & Forecast origin and horizon, reference target, discrepancy, and seed aggregation & What exactly does ``prediction failure'' measure? \\
Discovery protocol & Candidate library, query budget, scout rule, comparator pairing, and search score & How was the case found, and were budgets comparable? \\
Selected case & Stable identifier, typed values, search score, prediction risk, and activity status & Can the reported condition be reconstructed and inspected? \\
Held-out evidence & Disjoint seeds, control definition, and candidate-level aggregation & Does elevated prediction risk reproduce beyond the search measurements? \\
Neighborhood evidence & Valid edit rule, radius, anchor groups, and fresh evaluation seeds & Does elevated risk persist under nearby valid changes? \\
Task consequence & Controller, proposals, task metric, and its separation from prediction risk & Is a downstream outcome being measured, and what causal claim is supported? \\
\bottomrule
\end{tabularx}
\caption{Reporting fields for natural-input failure discovery. The fields separate discovery efficiency from evidence about the returned condition.}
\label{tab:supp-reporting}
\end{table*}

This separation also clarifies negative or mixed method results. A search policy can fail to outperform a comparator while still returning a valid high-risk case; conversely, a high search score does not by itself establish reproducibility or local persistence. For that reason, the main paper reports search, held-out point, and neighborhood results under their own protocols and uses distinct terms for prediction risk, search score, and task consequence.

\raggedbottom

\end{document}